\documentclass[10pt,twocolumn,letterpaper]{article}
\usepackage[pagenumbers]{cvpr}

\usepackage[T1]{fontenc}
\usepackage{array,tabularx,multirow,longtable}
\usepackage{algorithm,algorithmic}
\usepackage{microtype} 
\usepackage{placeins}
\makeatletter
\AtBeginDocument{%
  \hfuzz=0pt
  \typeout{FORMAT-AUDIT: textwidth=\the\textwidth; textheight=\the\textheight; columnsep=\the\columnsep}%
  \typeout{FORMAT-AUDIT: font=\f@family; size=\f@size; baseline=\the\baselineskip; parindent=\the\parindent}%
  \typeout{FORMAT-AUDIT: floatsep=\the\floatsep; textfloatsep=\the\textfloatsep; abovedisplayskip=\the\abovedisplayskip}%
}
\makeatother

\usepackage{needspace}

\usepackage{stfloats}
\usepackage{lineno}
\nolinenumbers

\definecolor{linkblue}{rgb}{0.21,0.49,0.74}
\usepackage[breaklinks,colorlinks,allcolors=linkblue,hypertexnames=false]{hyperref}
\hypersetup{pdftitle={Completion-Aware Compression for Progressive 3D Gaussian Splatting},pdfauthor={}}
\title{Completion-Aware Compression for Progressive 3D Gaussian Splatting}
\author{Reagan Koo \quad Yeong-Gyu Kim \quad Eun-Seok Ryu\\Sungkyunkwan University\\{\tt\small \{reagan9,kyk2798,esryu\}@skku.edu}}
\hypersetup{pdfauthor={Reagan Koo; Yeong-Gyu Kim; Eun-Seok Ryu}}

\begin{document}
\maketitle
\raggedbottom
\begin{abstract}
3D Gaussian Splatting (3DGS) enables high-quality novel-view rendering, but its large scene representations are costly to transmit. Pruning and compression reduce file sizes, yet progressive delivery also requires usable reconstructions before the full scene arrives. Existing progressive representations provide partial scenes, but level-based updates are limited by predefined coding boundaries. This paper presents a progressive 3DGS codec that balances attribute fidelity with the image benefit of making Gaussians available earlier. Its single compact stream supports partial 3D reconstruction and adds newly decoded Gaussians as more bytes arrive. The stream supports fine-grained updates at arbitrary received-byte budgets, without separate encodings or later attribute refinement. At matched complete-file sizes and received bytes, completion-aware allocation improves 1/8 PSNR over attribute-only allocation by 0.48--1.43 dB. The proposed codec compresses original 3DGS representations by 62--88 times while retaining near-reference Full PSNR. Under the evaluated 50-Mbps delivery model, partial reconstruction becomes available in approximately 0.2~s.
\end{abstract}

\section{Introduction}
\begingroup
\setlength{\parskip}{0pt}
3D Gaussian Splatting (3DGS) enables high-quality novel-view rendering \cite{kerbl2023}, but its explicit scene representations often require hundreds of megabytes of storage. Transferring these files can delay access to a scene. Pruning reduces the number of Gaussians, and compression reduces the storage required by those retained \cite{gaussianspa,compgs,hacpp}. Reducing file size is only part of progressive delivery: the receiver must also reconstruct a usable partial 3D scene as soon as sufficient data arrive. Only fully decoded Gaussians can contribute to the rendering, which is updated as more bytes arrive.

\begin{figure}[!t]
\centering
\includegraphics[width=\columnwidth]{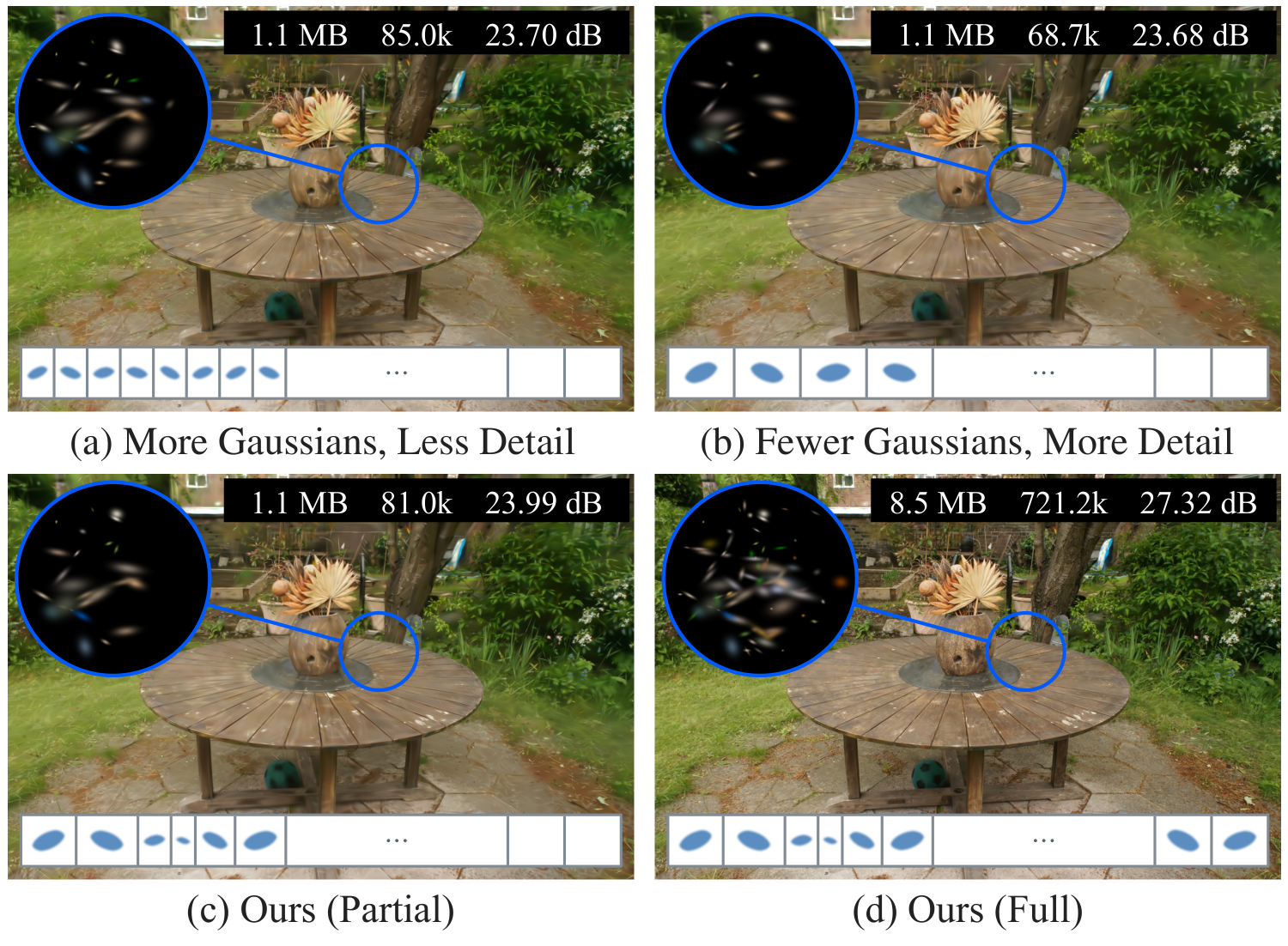}
\renewcommand{\thefigure}{1}\caption{Balancing content and detail at matched complete-file size. Panels (a--c) show Count-aware, Attribute-only, and Ours at equal received bytes; (d) completes the same Ours stream. Badges report received MB, decoded Gaussian count, and PSNR.}\label{fig:1}\end{figure}

Level-of-detail (LOD) and progressive representations organize scene content for different rendering or transmission budgets \cite{flexgs,octreegs,matryoshka,gode,pcgs,progs}. Selecting an LOD subset from an available representation differs from reconstructing a scene during transmission, where future data remain inaccessible. In a layered stream, further bytes may not expose a new reconstruction until the next coding unit is complete. To support finer updates, the proposed codec encodes each retained Gaussian as a complete record within one causal stream. Each completed Gaussian can then be added without waiting for a predefined layer boundary. This granularity couples attribute precision to subsequent availability: more bits improve an early Gaussian's fidelity but can postpone later Gaussians at the same received-byte budget. Fewer bits can reveal additional content earlier, while the precision loss persists at Full because decoded attributes are not refined later.

A favorable Gaussian order or a small final file alone cannot guarantee high-quality partial reconstructions. Equal bit savings can have different visual effects: shortening one record may make Gaussians that fill missing scene regions available sooner, while shortening another may advance Gaussians with little visible impact. Completed count alone cannot capture this difference. Each cutoff observes the same encoding, so a precision choice made for early reception also constrains later reconstructions. The challenge is to allocate bits so that partial scenes look good as bytes arrive while the complete scene retains high quality.

Figure 1 illustrates this allocation trade-off. At identical received bytes, completion-aware allocation (c) achieves higher PSNR than Count-aware (a) while decoding fewer Gaussians, and higher PSNR than Attribute-only (b) while decoding more. Further bytes extend the same stream to Full (d). This example exposes the allocation trade-off: additional Gaussians provide scene content, but their image value depends on visibility, overlap, and the precision of the decoded attributes. Table 3 tests this contrast across all three datasets using a fixed prepared representation and size-matched controls.

This paper presents a progressive 3DGS codec that balances the quality of decoded Gaussian attributes with the image benefit of making subsequent Gaussians available earlier. Scene preparation determines the retained Gaussians, spherical-harmonic (SH) degrees, transmission order, and scale--opacity correction (S+O) coefficients. With these choices fixed, completion-aware allocation optimizes quantization and entropy-coding settings. Quantization changes both attribute error and record length, while an entropy context changes the length of the same quantized values. A reception-weighted objective evaluates how these changes affect partial and Full images. Rendering sensitivities approximate the image contribution of shifted completion boundaries, allowing the allocator to compare attribute distortion, completion cost, and delivered rate without encoding and rendering every candidate. Controlled experiments at matched complete-file sizes and identical partial bytes show higher early PSNR and lower BD-rate than attribute-only and Count-aware allocation across all three datasets, despite Count-aware completing more Gaussians. Intermediate cutoffs test the same encoding between the principal reporting budgets. A persistent decoder retains its state across arriving chunks and adds complete Gaussians without replaying earlier bytes. Thus, the optimized stream supports both early scene access and the complete reconstruction without separate encodings for different reception budgets.

The contributions are:

\begin{itemize}
\item A progressive rate--distortion objective that jointly evaluates Gaussian attribute fidelity and the scene content available at each received-byte budget.
\item An image-based Gaussian completion cost and practical allocator that estimate how an early record's coded length affects downstream reconstruction.
\item A single causal codec with persistent incremental decoding, validated through fixed-representation allocation controls and crossed reconstructions that examine interactions with partial-scene correction.
\end{itemize}

\endgroup
\section{Related Work}

\textbf{Pruning and simplification.} Image-guided pruning and compact scene construction reduce the number and complexity of Gaussians that must be delivered \cite{compact3dgs,lightgaussian,compressed3dgs,pup,gaussianspa,maskgaussian,omg,megs2}. LightGaussian combines visibility across training views, opacity, and volume in a global significance score, then fine-tunes the survivors \cite{lightgaussian}. PUP uses second-order reconstruction sensitivity to spatial parameters and alternates pruning with refinement \cite{pup}. MaskGaussian learns existence probabilities and propagates gradients to temporarily unused Gaussians, allowing importance to evolve during optimization \cite{maskgaussian}. These methods use rendered contributions or reconstruction sensitivity to distinguish Gaussians, rather than treating all removals equally. Subsequent optimization lets the remaining Gaussians compensate for deleted content. Pruning determines which primitives and appearance components remain in the final representation; compression determines how efficiently that retained information is encoded.

\textbf{Gaussian compression.} Attribute quantization, codebooks, transforms, and image-codec-friendly layouts reduce the bits required by retained Gaussians \cite{compgs,mesongs,flexgaussian,kissgs,featureplanes2025}. Learned and causal entropy models exploit remaining statistical dependencies \cite{hac,hacpp,contextgs,fcgs,wang2026racergs,yu2026practicalcompression3dgaussian}. RDO-Gaussian and SizeGS explicitly balance reconstruction quality against rate or a size constraint \cite{rdo,sizegs}. Attribute groups can receive different quantization precision, allowing coding to preserve the parameters that most affect the rendered image. Rendering sensitivities connect attribute errors and primitive removal to image distortion under visibility and overlap \cite{compressed3dgs,pup}. These methods determine how efficiently the retained information is stored and how accurately it is reconstructed at encoded operating points. In a partially received file, coded length additionally determines how much scene information is accessible.

\textbf{Progressive representations and LOD.} LOD methods adapt Gaussian subsets to rendering requirements \cite{flexgs,octreegs,flod}, while continuous orders, learned hierarchies, and nested subsets vary scene content with the budget \cite{clod,ldh,matryoshka,progs_wacv}. GoDe provides reusable Gaussian hierarchies \cite{gode}; PCGS adds anchors and refines earlier attributes at predefined levels \cite{pcgs}; ProGS uses parent-closed octree prefixes with parent-causal coding \cite{progs}. LapisGS combines additive layers with opacity adjustment \cite{lapis}, and SHIN shares stage budgets between Gaussian additions and SH bands \cite{koo2026shinscalablehierarchyincremental}. SCAR-GS studies progressive residual-vector refinement, and 4DGCPro extends hierarchical compression to volumetric video \cite{scargs,4dgcpro}. Existing methods provide compact Gaussian representations and progressively usable scene structures. This motivates an allocation objective that considers both attribute fidelity and decoded scene content across intermediate received-byte budgets.

\begin{figure*}[!t]
\centering
\includegraphics[width=\textwidth]{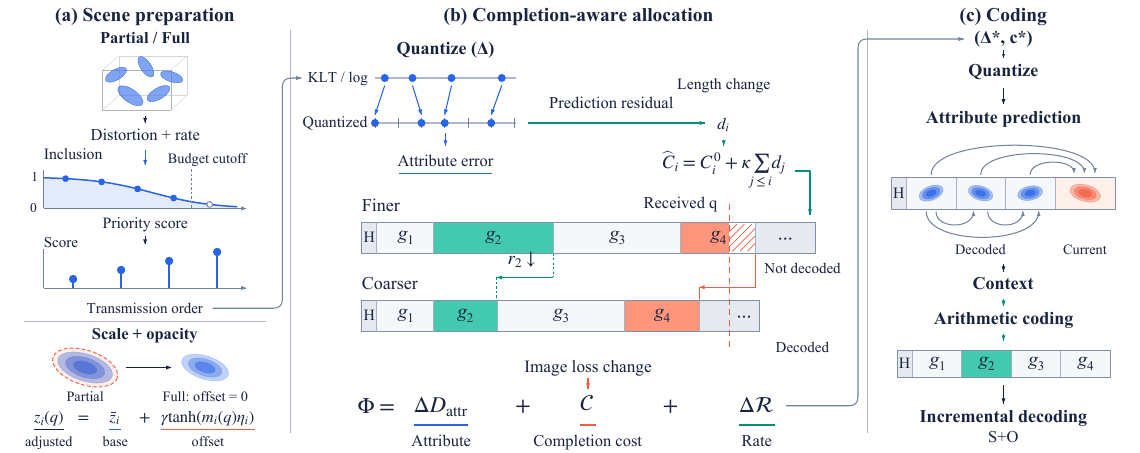}
\renewcommand{\thefigure}{2}\caption{Completion-aware compression. (a) Partial and Full image losses learn Gaussian retention, SH degrees, transmission priorities, and S+O coefficients. The prepared representation is then fixed. (b) Allocation predicts how a change in an early record's length shifts downstream completion and evaluates the resulting image contribution against attribute distortion and rate. (c) Causal prediction and arithmetic coding produce one ordered stream for persistent incremental decoding.}\label{fig:2}\end{figure*}

\section{Method}

Separating scene preparation from coding allocation allows precision decisions to be evaluated on a common Gaussian representation. Preparation learns Gaussian retention, SH degrees, transmission order, and partial-scene correction. With these choices fixed, allocation optimizes attribute precision and entropy-coding settings before causal encoding produces a single incremental stream (Figure 2).

\subsection{Preliminaries and Problem Setup}

3DGS forms images by depth-ordered alpha compositing of Gaussians with position, covariance, opacity, and SH color \cite{kerbl2023}. Each retained Gaussian is encoded once as a Gaussian record containing its required attributes; it becomes renderable only after that record is fully decoded. For a given transmission order, let stream $S$ have header length $H$ and record lengths $r_i$, in bits. Completion positions $C_i$ and the set renderable after receiving $q$ bits are
\begin{equation}
C_i=H+\sum_{j\le i}r_j,\qquad
\mathcal A_S(q)=\{i:C_i\le q\}.
\tag{1}
\end{equation}

The image $I_v(S,q)$ renders this completed prefix at view $v$. Changing one record's length can move subsequent Gaussians across the same cutoff $q$, changing the available scene content even when transmission order is fixed.

\subsection{Preparing the Ordered Representation}

The first decoded Gaussian subsets must approximate scene structure and appearance before the full stream arrives. Preparation jointly learns Gaussian retention, SH degrees, transmission priorities, and scale--opacity correction from partial and Full reconstruction losses with rate regularization.

\textbf{Learning the transmission order.} A Gaussian set that reconstructs Full well may yield poor early images in an arbitrary order. Priorities are therefore optimized at multiple reception budgets. Log-uniform sampling balances equal proportional byte increases, while log-MSE values proportional image-error reductions. A separate Full term evaluates complete-reconstruction quality alongside these early gains.

To learn priorities from partial images, training uses a soft gate to approximate discrete Gaussian inclusion. For priorities $p_i$ and estimated lengths $\widehat r_i$, threshold $\tau_b$ satisfies payload budget $B_{\mathrm{record}}(b)$ at fraction $b$, with sharpness $\beta$:
\begin{equation}
\begin{aligned}
a_i(b)&=\operatorname{sigmoid}\!\big(\beta(\tau_b-p_i)\big),\\
\sum_i\widehat r_i a_i(b)&=B_{\mathrm{record}}(b).
\end{aligned}
\tag{2}
\end{equation}

Forward rendering uses complete Gaussians; ascending priority defines the order $\pi=\operatorname{argsort}(p)$.

\textbf{Gaussian retention and SH degree.} Each Gaussian is omitted or retained with degree 0--3 for its view-dependent appearance. Nested continuation gates keep lower SH shells active when higher ones are selected. Image cost and coded length finalize these states; the subsequent allocation determines quantization precision (\hyperref[supp:A1]{Supplement A.1}).

\textbf{Partial-scene correction.} Incomplete subsets may insufficiently cover the scene or reproduce accumulated opacity. S+O adjusts the scale and opacity of decoded Gaussians. For decoded base coordinate $\bar z_i$ and coefficient $\eta_i$, the offset depends on received length $q$, completion position $C_i$, and complete length $L$:
\begin{equation}
\begin{aligned}
m_i(q)&=\frac{\ln(L/q)}{\ln(L/C_i)},\qquad C_i\le q<L,\\
z_i(q)&=\bar z_i+\gamma\tanh\!\big(m_i(q)\eta_i\big),\quad m_i(L)=0.
\end{aligned}
\tag{3}
\end{equation}

Signed coefficients allow either adjustment direction, bounded by $\gamma$; the correction magnitude decreases to zero at Full. Preparation then fixes the retained Gaussians, base attributes, SH degrees, transmission order, and S+O coefficients as $\Theta$.

\begin{algorithm}[!t]
\caption{Completion-aware compression}\label{alg:main}
\setboolean{ALC@noend}{true}
\begin{algorithmic}[1]
\REQUIRE Fixed $\Theta$, seed $S_0$, density $p(b)$, legal sets $\mathcal V_j\ni p_j$
\STATE Fix $\{q_k,w_k\},M,\{g_k,h_k\}$; $\mathbf p\gets(\Delta_0,c_0)$
\FOR{$\mathcal B$ in $(\mathcal B_\Delta^1,\ldots,\mathcal B_\Delta^m,\mathcal B_c)$}
\STATE $t\gets0$
\REPEAT
\STATE $\mathbf p_{\mathrm{pass}}\gets\mathbf p$; $t\gets t+1$
\FOR{$j\in\mathcal B$}
\FOR{$v\in\mathcal V_j$}
\STATE $d_i\gets\ell_i(\mathbf p[j\gets v])-\ell_i(S_0)$ \COMMENT{ideal bits}
\STATE $(\widehat C,\widehat L)\gets$ Equation (6)
\STATE $s_v\gets\Phi(\mathbf p[j\gets v])$ using (7)
\ENDFOR
\STATE $p_j\gets\arg\min_{v\in\mathcal V_j}s_v$; keep current on ties
\ENDFOR
\UNTIL{$\mathbf p=\mathbf p_{\mathrm{pass}}$ or $(\mathcal B\ne\mathcal B_c\land t=2)$}
\ENDFOR
\RETURN $S^\star\gets\mathrm{Encode}(\Theta,\mathbf p)$
\end{algorithmic}
\end{algorithm}

\subsection{Completion-Aware Rate--Distortion Objective}

Allocation selects quantization steps $\Delta$ and entropy contexts $c$ while fixing $\Theta$. A larger step can shorten a record but changes its attributes; an entropy context changes the length of the same quantized values. Precision is shared by attribute component and four contiguous order intervals, with one context per component, limiting signaling overhead and defining the search coordinates (\hyperref[supp:A2]{Supplement A.2}). A shared step can shift completion beyond its own order interval, while the completion cost is evaluated per Gaussian.

A precision choice can improve one cutoff while degrading another or Full, so a single-cutoff objective is insufficient. An initial stream $S_0$ of length $L_0$ fixes byte-rounded requests $q(b)=8\lfloor bL_0/8\rfloor$; candidates share absolute bytes rather than fractions of their own lengths. Partial density $p(b)=1/(b\ln8)$ on $[1/8,1)$ weights equal logarithmic reception intervals equally, and Full has weight $\mu=1/2$. For candidate length $L$,
\begin{equation}
\begin{aligned}
\mathcal J(S)={}&(1-\mu)\mathbb E_b\!\left[\mathcal D\big(S,\min(q(b),L)\big)\right]\\
&+\mu\mathcal D(S,L)+\mathcal R(L),\\
\mathcal R(L)={}&\ln2\!\left[(1-\mu)\mathbb E_b[\min(q(b),L)]+\mu L\right].
\end{aligned}
\tag{4}
\end{equation}

The image term $\mathcal D(S,q)=\frac12\sum_v n_v\ln D_v(S,q)$ uses RGB MSE $D_v$ over $n_v$ color components; the $\ln2$ conversion puts delivered bits on the same scale. Candidates ending before a request contribute their Full reconstruction and actual length. Remaining partial states are integrated over reception (\hyperref[supp:B2]{Supplement B.2}).

To identify the effect of completion timing alone, hold the rendering of each ordered prefix fixed. Let $D_i$ be the loss after the first $i$ Gaussians are decoded; loss then changes only when another record completes. For partial received length $Q$ and survival probability $\bar F(x)=\Pr(Q\ge x)$,
\begin{equation}
\begin{aligned}
\mathbb E[D(Q)]&=D_0+\sum_i(D_i-D_{i-1})\bar F(C_i),\\
\frac{\partial\mathbb E[D(Q)]}{\partial r_j}
&=-\sum_{i\ge j}(D_i-D_{i-1})f_Q(C_i).
\end{aligned}
\tag{5}
\end{equation}

Changing $r_j$ shifts all subsequent $C_i$. Equal bit savings can therefore yield different image benefits, depending on which Gaussians cross a cutoff and their image-loss transitions $D_i-D_{i-1}$; a uniform count reward cannot distinguish these transitions. Equation (5) fixes attributes, the header, and prefix renderings, with continuous density $f_Q(x)=1/(x\ln8)$ within its support. The full objective recomputes S+O (\hyperref[supp:B2]{Supplement B.2}).

\begin{table*}[!t]
\centering
\captionsetup{type=table}
\small
\setlength{\tabcolsep}{2.5pt}\renewcommand{\arraystretch}{1.12}
\renewcommand{\thetable}{1}\caption{Partial and complete reconstruction against 3DGS \cite{kerbl2023}. Dataset means; PSNR is in dB and size in MB. The proposed codec uses one stream; 3DGS uses the complete PLY.}\label{tab:1}
\begin{tabular*}{\textwidth}{@{\extracolsep{\fill}}lrrrrrrrrrrrr@{}}
\toprule
&\multicolumn{4}{c}{Mip-NeRF360 \cite{mipnerf360}}&\multicolumn{4}{c}{Tanks \& Temples \cite{tanks}}&\multicolumn{4}{c}{Deep Blending \cite{deepblending}}\\
\cmidrule(lr){2-5}\cmidrule(lr){6-9}\cmidrule(lr){10-13}
Method / fraction & PSNR $\uparrow$ & SSIM $\uparrow$ & LPIPS $\downarrow$ & MB $\downarrow$ & PSNR $\uparrow$ & SSIM $\uparrow$ & LPIPS $\downarrow$ & MB $\downarrow$ & PSNR $\uparrow$ & SSIM $\uparrow$ & LPIPS $\downarrow$ & MB $\downarrow$\\\midrule
3DGS \cite{kerbl2023} & 27.80 & 0.8264 & 0.2015 & 640.96 & 23.49 & 0.8308 & 0.2350 & 449.01 & 29.67 & 0.9025 & 0.2413 & 608.86\\
\midrule
Ours, 1/8 & 24.92 & 0.6603 & 0.4098 & 0.91 & 21.62 & 0.7085 & 0.4067 & 0.90 & 28.26 & 0.8617 & 0.3412 & 0.91\\
Ours, 1/4 & 26.54 & 0.7310 & 0.3415 & 1.82 & 23.02 & 0.7619 & 0.3480 & 1.80 & 29.28 & 0.8848 & 0.3000 & 1.81\\
Ours, 1/2 & 27.29 & 0.7720 & 0.2888 & 3.64 & 23.48 & 0.7898 & 0.3054 & 3.61 & 29.58 & 0.8948 & 0.2752 & 3.63\\
Ours, Full & 27.76 & 0.8000 & 0.2446 & 7.28 & 23.64 & 0.8047 & 0.2739 & 7.21 & 29.68 & 0.8990 & 0.2567 & 7.26\\
\bottomrule
\end{tabular*}
\end{table*}

\subsection{Image-Aware Completion Cost}

Equation (5) identifies the image effect of shifted completion boundaries, but evaluating each candidate through actual encoding and partial rendering would be costly. The allocator therefore predicts boundary shifts from coded-symbol lengths and estimates their image effects using sensitivities computed from the reference stream.

Ideal length sums $-\log_2 P(s\mid\mathrm{ctx})$ over coded symbols under the adaptive entropy context. Let $d_i$ be the candidate-minus-reference length. The measured-to-ideal payload ratio $\kappa$ from $S_0$ maps this change to measured completion coordinates $C_i^0$:
\begin{equation}
\widehat C_i=C_i^0+\kappa\sum_{j\le i}d_j,\qquad
\widehat L=L_0+\kappa\sum_jd_j.
\tag{6}
\end{equation}

In Figure 2(b), shortening $g_2$ brings $g_4$ before cutoff $q$. Reception probability between old and new completion positions weights the changed inclusion. Adopted changes accumulate relative to $S_0$, with header changes handled separately.

Rendering sensitivities estimate the completion cost $\mathcal C$, the image effect of changes in the decoded Gaussian set, alongside attribute quantization error. For candidate and reference errors $e,e_0$, a Gauss--Newton metric $M$ aggregates partial and Full views. Normalized by the reference RGB MSE, $M$ weights attribute errors by their rendered effect, giving equal parameter errors different image costs. The inclusion gradient $g_{ki}$ measures the local loss response to changing Gaussian $i$'s availability at request $q_k$; the nonnegative Gauss--Newton curvature $h_{ki}$ measures its squared image response. Both are evaluated in the reconstruction at that request, where the available prefix determines transmittance and the color composited behind the Gaussian. Writing the inclusion change as $\delta a_{ki}=\mathbf1[\widehat C_i\le q_k]-\mathbf1[C_i^0\le q_k]$ gives the combined score
\begin{equation}
\begin{aligned}
\Phi={}&\underbrace{\tfrac12(e^TMe-e_0^TMe_0)}_{\Delta D_{\mathrm{attr}}}
+\underbrace{\mathcal R(\widehat L)-\mathcal R(L_0)}_{\Delta R}\\
&+\underbrace{(1-\mu)\sum_k w_k\sum_i
          \left(g_{ki}\delta a_{ki}+\tfrac12h_{ki}\delta a_{ki}^2\right)}_{\text{completion cost }\mathcal C}.
\end{aligned}
\tag{7}
\end{equation}

{\widowpenalty=10000
Inclusion changes of $+1$ and $-1$ add and remove image contributions; weights $w_k$ integrate their estimated loss changes over reception. Insertion and removal reverse the linear contribution but retain the quadratic term, so their predicted costs need not be symmetric. For a Gaussian not yet decoded at a reception cutoff, the encoder evaluates its inclusion derivative at zero inclusion using preceding transmittance, ungated alpha, and contrast with the color composited behind it (\hyperref[supp:B1]{Supplement B.1}). The three terms serve different roles: $\Delta D_{\mathrm{attr}}$ estimates the change in attribute quantization distortion, $\mathcal C$ values changed content availability, and $\Delta R$ accounts for delivered length. An entropy-context change can move boundaries without changing quantized base attributes; coarser quantization can instead advance completion while increasing attribute error.\par}

Six Gauss--Legendre nodes in log reception fraction approximate partial reception, with Full evaluated separately. Doubling the nodes changes the measured objective on captured configurations by at most 0.0372\%, checking integration resolution independently of the coding and image-response approximations (\hyperref[supp:B2]{Supplement B.2}).

\subsection{Optimization and Incremental Decoding}

Rendering sensitivities are computed once from the initial stream $S_0$. Each candidate is then scored using updated quantization errors and predicted completion positions, without repeatedly encoding and rendering the scene. Algorithm 1 searches precision blocks $\mathcal B_\Delta^a$ and then context coordinates $\mathcal B_c$, keeping current settings on ties. Quantization determines the symbol distributions used for context selection. The selected stream is encoded and evaluated with actual boundaries and recomputed S+O (\hyperref[supp:B3]{Supplement B.3}).

With coding settings fixed, the receiver maintains one causal arithmetic decoder across byte chunks (Figure 2(c)). It retains the interval, code register, input cursor, adaptive symbol counts, and spatial prediction context, so symbol processing continues from its interrupted state. A Gaussian becomes renderable only when all required attributes and its final symbol's renormalization bits are available. Pending records wait for further bytes while completed ones stay visible, without earlier bytes being replayed. S+O follows received length and actual completion positions, and decoded base attributes keep their assigned precision (\hyperref[supp:A2]{Supplement A.2}).

\setcounter{topnumber}{2}

\section{Experiments}

\subsection{Evaluation Setup}

The evaluation covers 13 scenes from Mip-NeRF360, Tanks \& Temples, and Deep Blending \cite{mipnerf360,tanks,deepblending}. Original 3DGS is the uncompressed Full reference; GoDe, PCGS, ProGS, and RDO-Gaussian are codec baselines \cite{kerbl2023,gode,pcgs,progs,rdo}. PCGS High-rate uses the authors' original settings. Low-rate uses stronger rate penalties to overlap the proposed stream's byte range, with one setting fixed across all scenes of each dataset and the native training schedule retained. The proposed method uses one encoded stream per scene for every reception budget. Allocation controls share its prepared representation and identical partial bytes. \hyperref[supp:A3]{Supplement A.3} specifies baseline files, levels, and common-range BD evaluation; \hyperref[supp:E]{Supplement E} gives delivery protocols.

\setcounter{dbltopnumber}{1}

\begin{figure*}[!t]
\centering
\includegraphics[width=\textwidth]{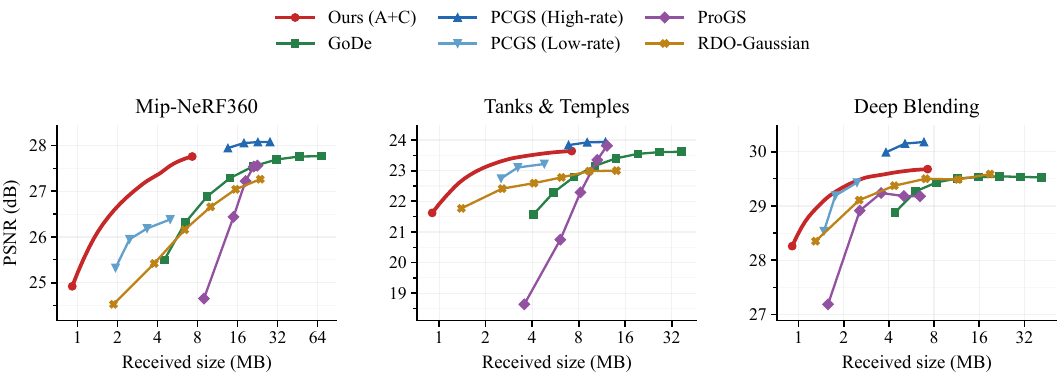}
\renewcommand{\thefigure}{3}\caption{Rate--distortion during reception on the three datasets \cite{mipnerf360,tanks,deepblending}, compared with GoDe, PCGS, ProGS, and RDO-Gaussian \cite{gode,pcgs,progs,rdo}.}\label{fig:3}

\centering

\end{figure*}

\begin{table}[!t]
\centering
\captionsetup{type=table}\small
\setlength{\tabcolsep}{1pt}\renewcommand{\arraystretch}{1.12}
\renewcommand{\thetable}{2}\caption{Common-range BD-PSNR (dB, $\uparrow$) and BD-rate (\%, $\downarrow$). PCGS-H/L denote High-/Low-rate PCGS \cite{pcgs}; RDO-GS denotes RDO-Gaussian. N/A indicates no valid common interval. Datasets follow \cite{mipnerf360,tanks,deepblending}.}\label{tab:2}
\begin{tabular*}{\columnwidth}{@{\extracolsep{\fill}}lrrrrrr@{}}
\toprule
&\multicolumn{2}{c}{Mip-NeRF360}&\multicolumn{2}{c}{Tanks \& Temples}&\multicolumn{2}{c}{Deep Blending}\\
\cmidrule(lr){2-3}\cmidrule(lr){4-5}\cmidrule(lr){6-7}
Baseline & dB & \% & dB & \% & dB & \%\\\midrule
GoDe \cite{gode} & +1.54 & -73.13 & +1.36 & -79.38 & +0.47 & -68.74\\
PCGS-H \cite{pcgs} & N/A & N/A & N/A & N/A & -0.49 & N/A\\
PCGS-L \cite{pcgs} & +0.82 & -25.05 & +0.36 & -34.81 & +0.10 & -18.67\\
ProGS \cite{progs} & N/A & -79.37 & +3.48 & -82.56 & +0.64 & -54.65\\
RDO-GS \cite{rdo} & +1.19 & -64.22 & +0.89 & -62.17 & +0.29 & -36.86\\
\bottomrule
\end{tabular*}
\end{table}

\begin{figure}[!t]
\centering
\includegraphics[width=\columnwidth]{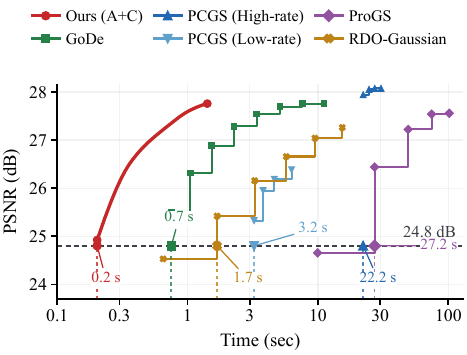}
\renewcommand{\thefigure}{4}\caption{Display-time PSNR at 50 Mbps on Mip-NeRF360 \cite{mipnerf360}, with baselines \cite{gode,pcgs,progs,rdo}. Dashed guides mark the 24.8 dB target and first qualifying displays; curves interpolate evaluated states.}\label{fig:4}
\end{figure}

At half reception, 3.61--3.64 MB yields PSNR within 0.51 dB of original 3DGS across all datasets, using 124--176 times fewer received bytes than the original complete PLY files (Table 1). This compares a received prefix with the full uncompressed reference; the complete compressed streams are 62--88 times smaller than those PLY files. Full PSNR remains near the reference, while lower SSIM and higher LPIPS indicate perceptual differences under the log-MSE objective.

\subsection{Reconstruction during Reception}

Over common observed intervals, the proposed stream gains 0.47--1.54 dB BD-PSNR over GoDe \cite{gode} with 68.74--79.38\% lower BD-rate, and 0.29--1.19 dB over RDO-Gaussian \cite{rdo} with 36.86--64.22\% lower BD-rate (Table 2). Against PCGS Low-rate \cite{pcgs}, it gains 0.10--0.82 dB with 18.67--34.81\% lower BD-rate across the three datasets. These common-range results assess compression efficiency. Separately, the proposed single stream renders completed Gaussians at intermediate received-byte positions, including budgets below the comparison codecs' first evaluated operating points (Figure 3).

\begin{figure*}[t]
\centering
\includegraphics[width=.94\textwidth]{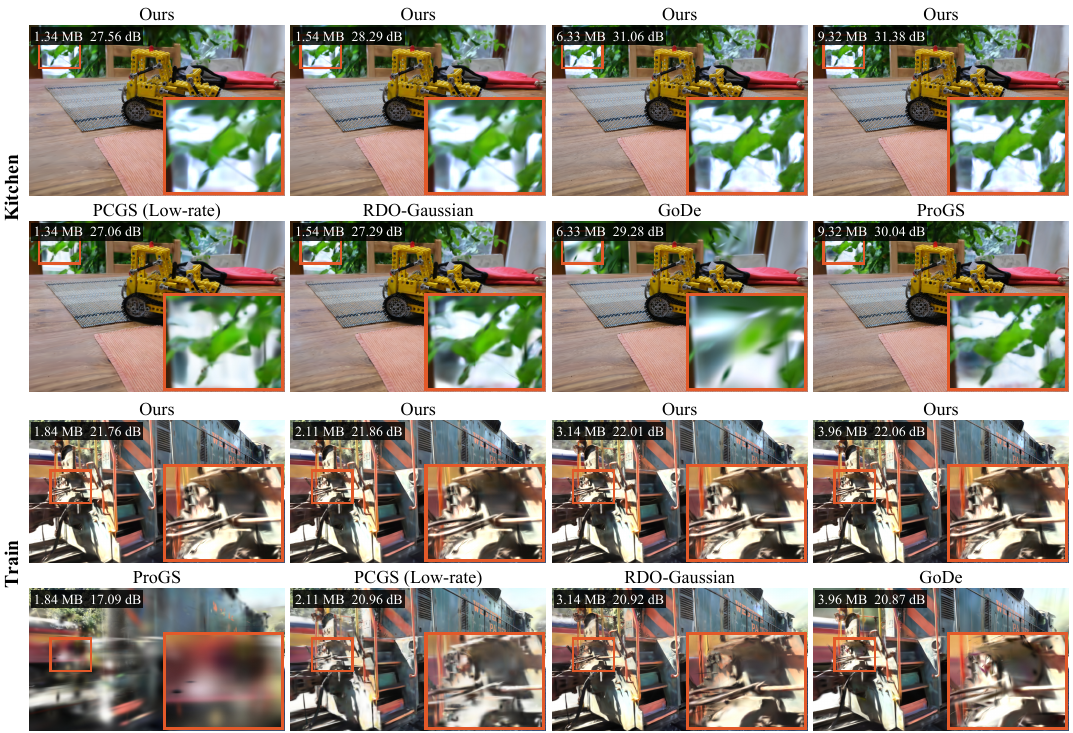}
\renewcommand{\thefigure}{5}\caption{Kitchen \cite{mipnerf360} and Train \cite{tanks} at matched bytes. All methods use the same camera and crop per scene; PSNR is the scene average. Baselines: PCGS, RDO-Gaussian, GoDe, and ProGS \cite{pcgs,rdo,gode,progs}.}\label{fig:5}
\end{figure*}

\Needspace{3\baselineskip}
Figure 4 compares display times at 50 Mbps by combining modeled byte arrivals with measured decoding, reconstruction, and rendering times. On Mip-NeRF360, the proposed stream reaches the 24.8 dB target at 0.2~s, compared with 0.7~s for GoDe, 1.7~s for RDO-Gaussian, 3.2~s for PCGS Low-rate, 22.2~s for PCGS High-rate, and 27.2~s for ProGS. Additional bandwidths, quality targets, and processing-only times are reported in \hyperref[supp:E]{Supplement E}.

Figure 5 decodes the same proposed stream at each baseline's byte budget, rendering complete Gaussians without reallocating precision. All methods use the same camera and crop within each scene; labels report scene-average test PSNR at the indicated bytes. Zooms highlight contours and repeated structures. \hyperref[supp:C3]{Supplement C.3} evaluates intermediate cutoffs, and \hyperref[fig:A3]{Figure A3} shows the proposed progression across all 13 scenes.

\setcounter{topnumber}{1}
\subsection{Ablation Studies}

To isolate the effect of completion-aware allocation within the codec, Table 3 compares $A+C$ against attribute-only ($A$) and Count-aware controls under a shared prepared representation. Their complete-file sizes agree within 0.5\%, and partial reconstructions use identical received bytes.

Figure 6 extends these controls and the tuned order-only schedule to 29 states from 1/8 to Full. Relative to attribute-only, A+C makes more Gaussians available and improves early PSNR. Gaps narrow toward Full as completion differences diminish, while quantization differences remain.

Count-aware completes more Gaussians at identical partial bytes, yet A+C achieves higher mean early PSNR and lower BD-rate on every dataset. Completion and precision must therefore be evaluated together (\hyperref[supp:C1]{Supplement C.1}).

Uniform and interval-averaged completion controls retain the search but average image-derived sensitivities, removing Gaussian-specific variation. A+C improves their 1/8 PSNR by 0.51--1.56 dB and lowers BD-rate (\hyperref[tab:S3]{Table S3}). The tuned order-only schedule also falls short at early reception, indicating that averaged completion values and position-based precision schedules do not reproduce the gains of reception-dependent image sensitivity.

\begin{figure*}[!t]
\begingroup
\captionsetup{type=figure}
\centering
\includegraphics[width=.94\textwidth]{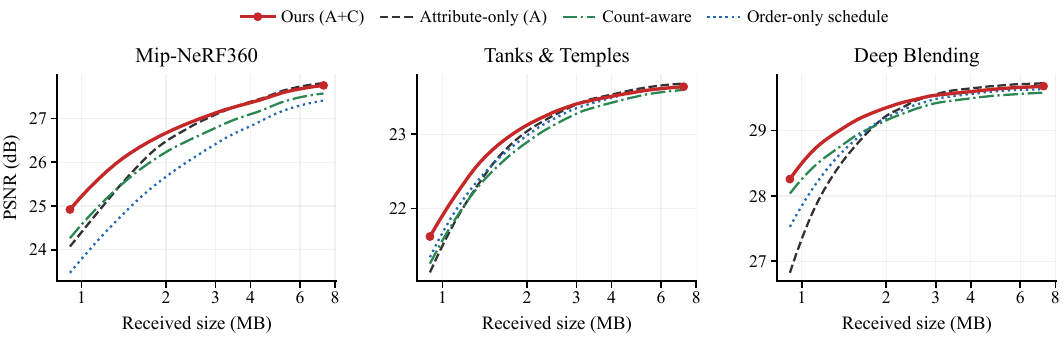}
\renewcommand{\thefigure}{6}\caption{Dense reception quality on the three datasets \cite{mipnerf360,tanks,deepblending}. The four allocations use 29 decoded states at common partial bytes; Full uses each complete file. Tables 3 and S3 report BD-rates.}\label{fig:6}
\par\endgroup\par
\end{figure*}

\subsection{Reconstruction Component Analysis}

Table 4 examines decoded Gaussian content, attribute precision, and S+O using a separate $A$/$A+C$ pair optimized with the same rate weight. With S+O disabled, crossed reconstructions change either the completed Gaussian set or the decoded attributes while holding the other fixed. Enabling S+O with both sources fixed tests correction. These are conditional comparisons; they cannot be added as independent contributions or used to decompose the size-matched gains in Table 3.

Changing only the completed set improves partial reconstruction with S+O disabled, demonstrating the image contribution of additional Gaussians (Table 4). With attributes fixed and S+O enabled, changing the boundary map also improves PSNR, with gains of 1.50--2.90 dB at 1/8 and 0.43--0.79 dB at 1/4 (\hyperref[tab:S4]{Table S4}). In this latter test, a new boundary map changes both which Gaussians are renderable and the S+O offsets applied to them, even though the prepared coefficients are shared. At Full, the completed sets coincide and S+O vanishes, so only decoded attribute differences remain.

\begin{table}[t]
\begingroup
\centering\small
\captionsetup{type=table,skip=5pt}
\setlength{\tabcolsep}{2.3pt}\renewcommand{\arraystretch}{1.14}
\renewcommand{\thetable}{3}\caption{Size-matched allocation at equal partial bytes on the three datasets \cite{mipnerf360,tanks,deepblending}. $A$: Attribute-only; Count: Count-aware. Gains are Ours ($A+C$) $-$ Control; BD-rate vs. Control.}\label{tab:3}
\newsavebox{\AllocationTableBox}
\sbox{\AllocationTableBox}{\begin{tabular}{@{}llrrrrr@{}}
\toprule
\multirow{2}{*}{Dataset} & \multirow{2}{*}{Control} & \multicolumn{2}{c}{\shortstack{1/8 Gaussians\\(k)}} & \multicolumn{2}{c}{\shortstack{PSNR gain\\(dB, $\uparrow$)}} & \multirow{2}{*}{\shortstack{BD-rate\\(\%, $\downarrow$)}}\\
\cmidrule(lr){3-4}\cmidrule(lr){5-6}
 & & Control & Ours & 1/8 & Full & \\
\midrule
\multirow{2}{*}{Mip-NeRF360} & $A$ & 48.3 & 59.3 & +0.85 & -0.05 & -8.90\\
 & Count & 62.4 & 59.3 & +0.65 & +0.18 & -24.03\\
\addlinespace[2.5pt]
\multirow{2}{*}{Tanks \& Temples} & $A$ & 39.6 & 46.3 & +0.48 & -0.04 & -8.47\\
 & Count & 48.6 & 46.3 & +0.37 & +0.04 & -15.22\\
\addlinespace[2.5pt]
\multirow{2}{*}{Deep Blending} & $A$ & 40.8 & 57.3 & +1.43 & -0.05 & -17.92\\
 & Count & 60.3 & 57.3 & +0.22 & +0.10 & -18.68\\
\bottomrule\end{tabular}}
\typeout{TABLE3-NATURAL-WIDTH=\the\wd\AllocationTableBox; COLUMN-WIDTH=\the\columnwidth}
\usebox{\AllocationTableBox}
\par\endgroup
\end{table}

\begin{table}[t]
\centering
\begingroup
\captionsetup{type=table,skip=5pt}\small
\setlength{\tabcolsep}{1.5pt}\renewcommand{\arraystretch}{1.00}
\renewcommand{\thetable}{4}\caption{Conditional PSNR changes (dB) at equal bytes \cite{mipnerf360,tanks,deepblending}. Content: decoded Gaussians; Attr.: attribute precision.}\label{tab:4}
\begin{tabular*}{\linewidth}{@{\extracolsep{\fill}}llrrr@{}}
\toprule
Dataset & Recv. & Content & Attr. & S+O\\\midrule
\multirow{2}{*}{Mip-NeRF360} & 1/8 & +2.44 & -0.05 & +2.26\\
 & 1/4 & +1.33 & -0.09 & +1.22\\
\multirow{2}{*}{Tanks \& Temples} & 1/8 & +1.89 & -0.03 & +1.45\\
 & 1/4 & +0.76 & -0.05 & +0.57\\
\multirow{2}{*}{Deep Blending} & 1/8 & +3.91 & -0.05 & +1.28\\
 & 1/4 & +1.33 & -0.08 & +0.31\\
\bottomrule\end{tabular*}
\par\endgroup\par
\end{table}

\subsection{Measured Objective and Search Validation}

Stream-level evaluation computes Equation (4) using encoded completion boundaries and recomputed S+O. Relative to the initial stream, the final allocation reduces the measured objective on all three datasets: partial-reception distortion reductions outweigh the remaining Full-image penalties (\hyperref[tab:S5]{Table S5}).

Candidate validation also tests whether predicted completion positions are accurate enough to select beneficial coding updates. The test covers 56 encoded configurations from 18 captured search states in Bonsai, Train, and DrJohnson, including rejected alternatives. Near reception cutoffs, mean absolute completion-position errors are 123.49 bytes for precision changes and 25.52 bytes for context changes (\hyperref[tab:S2]{Table S2}). Every adopted precision update in these captured states reduces the measured objective; \hyperref[tab:S2a]{Table S2a} compares selected updates with the best actually encoded alternatives within each state. These encoded-candidate evaluations support the reliability of the approximate allocation decisions in the tested search states.

\section{Conclusion}

3D Gaussian Splatting enables high-quality novel-view rendering, but its large scene representations remain costly to transmit. This work presents a compact progressive codec that balances decoded attribute fidelity with the image benefit of making subsequent Gaussians renderable earlier. Completion-aware allocation improves early reconstruction and reduces BD-rate relative to precision-focused and count-focused alternatives at matched file sizes and received bytes. Across three datasets, complete streams occupy only 7.2--7.3 MB while retaining near-reference 3DGS PSNR. The same causal stream supports fine-grained reconstruction throughout reception, adding completed Gaussians without separate encodings for different budgets or subsequent attribute refinement. These results demonstrate that strong 3DGS compression and progressive delivery can be achieved within a single transmitted representation.

\label{maintext:end}
\raggedbottom

\clearpage
\setcounter{section}{0}
\pdfbookmark[0]{Supplementary Material}{supplementary}
\maketitlesupplementary
\raggedbottom
\bibcite{mipnerf360}{{1}{2022}{{Barron et~al.}}{{Barron, Mildenhall, Verbin, Srinivasan, and Hedman}}}
\bibcite{megs2}{{2}{2026{}}{{Chen et~al.}}{{Chen, Chen, Zou, Huang, Wang, Liu, Sun, and Wang}}}
\bibcite{fcgs}{{3}{2025{}}{{Chen et~al.}}{{Chen, Wu, Li, Lin, Harandi, and Cai}}}
\bibcite{hac}{{4}{2025{}}{{Chen et~al.}}{{Chen, Wu, Lin, Harandi, and Cai}}}
\bibcite{hacpp}{{5}{2025{}}{{Chen et~al.}}{{Chen, Wu, Lin, Harandi, and Cai}}}
\bibcite{pcgs}{{6}{2026{}}{{Chen et~al.}}{{Chen, Li, Wu, Lin, Harandi, and Cai}}}
\bibcite{gode}{{7}{2025}{{Di~Sario et~al.}}{{Di~Sario, Renzulli, Grangetto, Sugimoto, and Tartaglione}}}
\bibcite{lightgaussian}{{8}{2024}{{Fan et~al.}}{{Fan, Wang, Wen, Zhu, Xu, and Wang}}}
\bibcite{matryoshka}{{9}{2026}{{Guo et~al.}}{{Guo, Zhang, Aktas, Fogarty, Hu, Aslan, Li, Baykal, Miao, Bengtson, Zhou, Xia, Vasconcelos, and Oztireli}}}
\bibcite{pup}{{10}{2025}{{Hanson et~al.}}{{Hanson, Tu, Singla, Jayawardhana, Zwicker, and Goldstein}}}
\bibcite{deepblending}{{11}{2018}{{Hedman et~al.}}{{Hedman, Philip, Price, Frahm, Drettakis, and Brostow}}}
\bibcite{kerbl2023}{{12}{2023}{{Kerbl et~al.}}{{Kerbl, Kopanas, Leimk{\"u}hler, and Drettakis}}}
\bibcite{tanks}{{13}{2017}{{Knapitsch et~al.}}{{Knapitsch, Park, Zhou, and Koltun}}}
\bibcite{koo2026shinscalablehierarchyincremental}{{14}{2026}{{Koo et~al.}}{{Koo, Kim, and Ryu}}}
\bibcite{compact3dgs}{{15}{2024}{{Lee et~al.}}{{Lee, Rho, Sun, Ko, and Park}}}
\bibcite{omg}{{16}{2025{}}{{Lee et~al.}}{{Lee, Ko, and Park}}}
\bibcite{featureplanes2025}{{17}{2025{}}{{Lee et~al.}}{{Lee, Shu, Sanchez, Schierl, and Hellge}}}
\bibcite{flexgs}{{18}{2025{}}{{Liu et~al.}}{{Liu, Wang, Li, Cai, Wang, Li, Molchanov, Wang, and Wang}}}
\bibcite{maskgaussian}{{19}{2025{}}{{Liu et~al.}}{{Liu, Zhong, Zhan, Xu, and Sun}}}
\bibcite{clod}{{20}{2025}{{Milef et~al.}}{{Milef, Seyb, Keeler, Nguyen-Phuoc, Bo{\v z}i{\v c}, Kondguli, and Marshall}}}
\bibcite{kissgs}{{21}{2026}{{Morgenstern et~al.}}{{Morgenstern, Branschke, Fleischmann, Szatmari, Schlack, Barthel, Eisert, and Hilsmann}}}
\bibcite{compgs}{{22}{2025}{{Navaneet et~al.}}{{Navaneet, Pourahmadi~Meibodi, Abbasi~Koohpayegani, and Pirsiavash}}}
\bibcite{compressed3dgs}{{23}{2024}{{Niedermayr et~al.}}{{Niedermayr, Stumpfegger, and Westermann}}}
\bibcite{ldh}{{24}{2026}{{Pan et~al.}}{{Pan, Zhou, and Zha}}}
\bibcite{octreegs}{{25}{2025}{{Ren et~al.}}{{Ren, Jiang, Lu, Yu, Xu, Ni, and Dai}}}
\bibcite{scargs}{{26}{2026}{{Revilla et~al.}}{{Revilla, Suresh, Tsang, and Bhojan}}}
\bibcite{flod}{{27}{2025}{{Seo et~al.}}{{Seo, Choi, Son, and Uh}}}
\bibcite{lapis}{{28}{2025}{{Shi et~al.}}{{Shi, Morin, Gasparini, and Ooi}}}
\bibcite{progs}{{29}{2026}{{Tang et~al.}}{{Tang, Liu, Jiao, Zhang, Yang, Hou, and Wang}}}
\bibcite{flexgaussian}{{30}{2025}{{Tian et~al.}}{{Tian, Gao, Xianyu, Cui, and Zhang}}}
\bibcite{rdo}{{31}{2025}{{Wang et~al.}}{{Wang, Zhu, He, Feng, Deng, Bian, and Chen}}}
\bibcite{wang2026racergs}{{32}{2026}{{Wang et~al.}}{{Wang, Zeng, and Fu}}}
\bibcite{contextgs}{{33}{2024}{{Wang et~al.}}{{Wang, Li, Guo, Yang, Kot, and Wen}}}
\bibcite{sizegs}{{34}{2025{}}{{Xie et~al.}}{{Xie, Liu, Zhang, Ge, Pan, Tang, Bai, Zhang, Fan, and Wang}}}
\bibcite{mesongs}{{35}{2025{}}{{Xie et~al.}}{{Xie, Zhang, Tang, Bai, Lu, Ge, and Wang}}}
\bibcite{yu2026practicalcompression3dgaussian}{{36}{2026}{{Yu et~al.}}{{Yu, Chen, Song, Qin, Zhang, Wang, and Guo}}}
\bibcite{gaussianspa}{{37}{2025}{{Zhang et~al.}}{{Zhang, Jia, Niu, and Yin}}}
\bibcite{4dgcpro}{{38}{2025}{{Zheng et~al.}}{{Zheng, Wu, Zhong, Tian, Cao, Xu, Yao, Zhang, Hu, and Zhang}}}
\bibcite{progs_wacv}{{39}{2025}{{Zoomers et~al.}}{{Zoomers, Wijnants, Molenaers, Vanherck, Put, Jorissen, and Michiels}}}

\renewcommand{\thesection}{\Alph{section}}
\newcounter{settingssaved}
This supplementary material details scene preparation and causal coding in Section A, followed by the image-cost approximation and allocation validation in Section B. Sections C--E present controlled allocation analyses, per-scene reconstructions, and decoding/display latency, respectively.

\section{Implementation and Evaluation}\label{supp:A}

\subsection{Scene Preparation}\label{supp:A1}

\textbf{Preparing useful partial scenes.} Main Section 3.2 introduces scene preparation for useful partial reconstructions and high Full quality. The preparation loss evaluates both reconstruction goals together with delivered rate: for view $v$ and reception fraction $b$, RGB MSE $D_v(b)$ and delivered-rate estimate $\widehat R_{\mathrm{del}}(b)$ define
\begin{equation}
\begin{aligned}
\mathcal L_{\mathrm{prep}}&=\mathbb E_{v,b}\big[\ln\max(D_v(b),10^{-8})
    +\lambda_{\mathrm{prep}}\widehat R_{\mathrm{del}}(b)\big],\\
\lambda_{\mathrm{prep}}&=\frac{2\ln2}{N_{\mathrm{RGB}}},\qquad
N_{\mathrm{RGB}}=\sum_v3H_vW_v.
\end{aligned}
\tag{S1}
\end{equation}

Log-MSE evaluates proportional error reductions at both partial and Full reception. With shuffled cameras and $u\sim U[0,1]$, $b=\min(8^{2u-1},1)$ assigns half the mass to Full and half uniformly in log reception fraction. Estimated rate combines initial core/SH lengths with current degree-symbol costs.

\textbf{Joint learning.} Starting from a densification-complete checkpoint, this objective jointly optimizes Gaussian retention, SH degrees, base attributes, transmission priorities, and S+O. Gaussian omission and SH-degree selection (0--3) are made trainable with four continuation variables; their nested products retain lower SH shells whenever a higher shell is selected:
\begin{equation}
s_{i\ell}=\prod_{j=0}^{\ell}u_{ij}.
\tag{S2}
\end{equation}

These projected variables use straight-through gradients and thresholding. Alongside retention and SH degree, transmission priorities determine which learned Gaussians enter early subsets. Priorities are therefore optimized from partial images through a budget-constrained inclusion gate, defined by estimated lengths $\widehat r_i$ and priorities $p_i$:
\begin{equation}
a_i=\operatorname{sigmoid}\big(\beta(\tau-p_i)\big),\qquad
\sum_i \widehat r_i a_i=B_{\mathrm{record}}.
\tag{S3}
\end{equation}

Forward images use complete Gaussians; Equation (S3) supplies priority gradients. Ascending priority determines the final transmission order.

\textbf{State selection and refinement.} After joint learning, the gate-based choices are converted into discrete omission or degree states. Each state $z$ is selected by $D_i(z)+\ln2\,\mathbb E[b]R_i(z)$ using a fixed image metric, six log-budget nodes, and Full mass. An encoding provides measured rates for one further selection pass.

With retention and SH states finalized, base attributes and SH degrees remain fixed while transmission priorities and S+O are refined using partial and Full image losses. This final stage improves the early subsets through their order and through scale--opacity adjustments that compensate for insufficient spatial coverage and opacity accumulation.

\textbf{S+O parameterization.} Both learning stages apply $\gamma\tanh(\tilde a_{i,p}+m_i\tilde a_{i,m})$ to coordinate $z_i$, with quantized coefficients denoted by tildes and normalized completion fraction $c_i$. Joint learning uses $m_i=\operatorname{clip}((1-b)/(1-c_i),0,1)$, and final refinement uses $m_i=\operatorname{clip}(\ln b/\ln c_i,0,1)$. The constant Full value is absorbed as $z_i^{\mathrm{base}}=z_i+\gamma\tanh(\tilde a_{i,p})$, preserving the Full endpoint before base-attribute quantization. The remaining partial coefficient $\tilde a_{i,m}$ is applied during reception as defined in Equation (S5). The representation is fixed for allocation; Table S0a lists settings.

\setcounter{settingssaved}{\value{table}}
\begin{table}[!ht]
\centering\footnotesize
\setlength{\tabcolsep}{3pt}\renewcommand{\arraystretch}{1.04}
\renewcommand{\thetable}{S0a}\caption{Scene-preparation settings. Joint learning updates attributes and transmission variables; final refinement fixes attributes and SH degrees.}\label{tab:S13}
\begin{tabularx}{\columnwidth}{@{}lX@{}}\toprule
Setting & Value\\\midrule
Initial checkpoint & Densification complete, 15K\\
Updates (joint / final) & 25K / 5K\\
SH-gate optimizer / LR & Projected Adam, $1.25\times10^{-4}$\\
Continuation range / threshold & $[0,1]$ / $1/2$\\
Gate sharpness (joint / final) & $\beta=24$ / $64$\\
Threshold gradient (joint / final) & Held fixed / implicit\\
Final optimizer & Adam\\
Final priority learning rate & $10^{-3}$\\
Final S+O learning rate & $5\times10^{-4}$\\
Final threshold solver & 32 bisections\\
\bottomrule\end{tabularx}
\end{table}
\setcounter{table}{\value{settingssaved}}

\Needspace{5\baselineskip}
\subsection{Coding Coordinates and Partial Correction}\label{supp:A2}

\textbf{Coordinates for precision allocation.} The prepared attributes are next expressed in coding coordinates that separate precision for Gaussian size and shape. In the matrix-log domain, covariance $\Sigma$ and opacity logit $o$ define log volume $v$, shape $B$, and log optical mass $u$:
\begingroup\predisplaypenalty=0
\begin{equation}
\begin{aligned}
v&=\tfrac12\ln\det\Sigma,\quad B=\ln\Sigma-\tfrac23vI,\\
u&=v+\ln\operatorname{softplus}(o).
\end{aligned}
\tag{S4}
\end{equation}
\endgroup

Removing the isotropic part of $\ln\Sigma$ leaves $B$ to describe orientation and relative axis lengths. This symmetric, trace-free matrix has five independent coordinates, encoded in a Frobenius-orthonormal basis so that Euclidean coefficient error equals the Frobenius error in $B$. Decoding reconstructs the positive-definite covariance $\Sigma=\exp(B+2vI/3)$ and inverts softplus to recover opacity. To reduce correlations before quantization, a shared KLT is applied to each SH shell. Table S0b lists the coding settings.

\setcounter{settingssaved}{\value{table}}
\begin{table}[!ht]
\centering\footnotesize
\setlength{\tabcolsep}{3pt}\renewcommand{\arraystretch}{1.18}
\renewcommand{\thetable}{S0b}\caption{Coding and allocation settings. S+O opacity amplitude applies to opacity logits. All intervals refer to the fixed transmission order.}\label{tab:S14}
\begin{tabularx}{\columnwidth}{@{}lX@{}}\toprule
Setting & Value\\\midrule
Position precision & 18-bit grid\\
Intrinsic base step & $1/128$\\
Correction base step & $2/127$\\
Step multipliers & $\{1,2,4,\ldots,512\}$\\
Order intervals & \shortstack[l]{$[0,1/8],\ [1/8,1/4]$\\$[1/4,1/2],\ [1/2,1]$}\\
S+O amplitude (scale / opacity) & $0.5$ / $2$\\
Precision blocks & Stability or at most two sweeps\\
Context search & Until stable\\
Ties & Retain current setting\\
\bottomrule\end{tabularx}
\end{table}
\setcounter{table}{\value{settingssaved}}

\textbf{Correction during reception.} Alongside the base attributes, each record carries the remaining partial-correction coefficients from A.1. For received size $q$, completion position $C_i$, and full length $L$, decoding applies the log-scale or opacity-logit correction as
\begin{equation}
\begin{aligned}
m_i(q)&=\frac{\ln(L/\min(q,L))}{\ln(L/C_i)},\\
z_i(q)&=\bar z_i+\gamma\tanh(m_i(q)\eta_i),\quad m_i(L)=0.
\end{aligned}
\tag{S5}
\end{equation}

Here $\bar z_i$ includes the absorbed Full correction and $\eta_i=(2/127)t_i$ is reconstructed from stored $t_i$. Separate coefficients adjust scale and opacity, with one scale coefficient shared across axes; signs determine direction. Clipping the ratio to $[0,1]$ with guarded denominators ensures zero correction at Full, including records completing there.

\textbf{Causal prediction and entropy contexts.} Each quantized attribute component is predicted from previously reconstructed records carrying it, indexed in Morton order. Prediction chooses the current code's predecessor or successor by squared position distance, with earlier record index breaking ties. An online affine fit uses the selected value, clips slope to $[-1,1]$, and falls back to a zero-initialized running mean. Cross-interval references are rescaled to the current quantization lattice.

The resulting residuals are coded under one of eight context models: a global model and every nonempty combination of SH degree, transmission-order interval, and a decoded-volume-dependent geometry-precision bin. Log volume excludes the last condition because it is coded before that bin becomes available. The residual model codes zero/nonzero status, sign, magnitude length, and mantissa bits separately.

Adaptive arithmetic coding uses Krichevsky--Trofimov probabilities \cite{arithmetic,kt}; decoder, adaptive-count, and prediction states persist across byte chunks. An interrupted symbol resumes without replay, and a Gaussian becomes renderable only after all attributes and the final symbol's renormalization bits arrive, while completed Gaussians remain visible.

Using these coordinates and causal coding rules, attribute-only allocation selects dyadic steps and contexts to encode the initial stream $S_0$, balancing rendered Gauss--Newton distortion against causal rate. Completion-aware allocation then adds the image effect of shifted completion positions to this starting configuration.

\FloatBarrier
\begin{table*}[t]
\centering\small
\setlength{\tabcolsep}{4pt}\renewcommand{\arraystretch}{1.08}
\renewcommand{\thetable}{S1}\caption{Per-scene offline preparation and compression cost \cite{mipnerf360,tanks,deepblending}. Stage times are dataset means in minutes; VRAM is the mean per-scene peak allocated memory.}\label{tab:offline_cost}
\begin{tabular*}{\textwidth}{@{\extracolsep{\fill}}lrrrrrrr@{}}
\toprule
Dataset & Initialization & \shortstack{Scene\\optimization} & \shortstack{Initial-stream\\coding} & Sensitivity & \shortstack{Allocation\\+ coding} & Total & \shortstack{Peak VRAM\\(GB)}\\\midrule
Mip-NeRF360 & 8.6 & 75.4 & 8.1 & 2.0 & 5.8 & 99.9 & 17.6\\
Tanks \& Temples & 5.4 & 72.1 & 8.3 & 2.1 & 6.2 & 94.0 & 19.3\\
Deep Blending & 8.2 & 71.9 & 7.7 & 1.1 & 6.0 & 94.9 & 16.8\\
\midrule
All scenes & 8.1 & 74.3 & 8.1 & 1.9 & 5.9 & 98.3 & 17.7\\
\bottomrule
\end{tabular*}
\end{table*}

\subsection{Evaluation}\label{supp:A3}

Prefixes are evaluated on held-out views at actual baseline byte sizes (main Figure 5); Full uses complete files. Controls share absolute partial budgets. Optimization requests are fixed from $S_0$; fractions use final \mbox{$A+C$} size. GoDe uses its hierarchy and quantization-aware refinement \cite{gode}; ProGS uses rate parameter 0.004, and RDO-Gaussian uses independent encodings \cite{progs,rdo}. BD-PSNR uses PCHIP interpolation with a monotonic quality envelope; BD-rate integrates log file size over common PSNR after identical dominated-point filtering \cite{bjontegaard,pchip}.

PCGS High-rate follows the authors' configuration \cite{pcgs}. Low-rate is retrained with stronger dataset-specific rate penalties to cover the proposed stream's byte range: $64\times$ the original weights on Mip-NeRF360 and $16\times$ on Tanks \& Temples and Deep Blending. Lambda-conditioning slopes are additionally scaled by $\min(1,16/\max\lambda)$ to stabilize exponential context activations. High-rate and Low-rate remain independently trained progressive streams.

\begin{table}[!b]
\captionsetup{type=table}
\centering

\centering\small
\setlength{\tabcolsep}{1.5pt}\renewcommand{\arraystretch}{.82}
\renewcommand{\thetable}{S1a}\caption{Persistent-receiver time (s), measured from session onset or the preceding reception cutoff \cite{mipnerf360,tanks,deepblending}.}\label{tab:S1}
\begin{tabular*}{\columnwidth}{@{\extracolsep{\fill}}lrrrrr@{}}
\toprule
Measure & 1/8 & 1/4 & 1/2 & 3/4 & Full\\\midrule
\multicolumn{6}{l}{\textbf{Mip-NeRF360}}\\
Cumulative & 0.20 & 0.36 & 0.70 & 1.05 & 1.42\\
Interval & 0.20 & 0.16 & 0.34 & 0.35 & 0.36\\
\addlinespace[3pt]
\multicolumn{6}{l}{\textbf{Tanks \& Temples}}\\
Cumulative & 0.18 & 0.35 & 0.69 & 1.03 & 1.39\\
Interval & 0.18 & 0.17 & 0.34 & 0.35 & 0.36\\
\addlinespace[3pt]
\multicolumn{6}{l}{\textbf{Deep Blending}}\\
Cumulative & 0.20 & 0.35 & 0.68 & 1.02 & 1.37\\
Interval & 0.20 & 0.15 & 0.33 & 0.34 & 0.35\\
\bottomrule
\end{tabular*}

\end{table}

\textbf{Offline preparation and compression cost.} Table S1 covers the densification-complete 15K checkpoint through the final stream, including intermediate encodings, validation, and file I/O. Stages separate rate-reference initialization, scene optimization, $S_0$ coding, sensitivity estimation, and final allocation and coding.

Offline costs use an NVIDIA RTX 6000 Ada GPU and an AMD EPYC 9124 CPU with 16 physical coding cores.

\subsection{Incremental Decoding Time}\label{supp:A4}

Table S1a reports cumulative processing since session onset and incremental processing between reception cutoffs. The receiver preserves decoder state across updates; Section E combines these measured services with byte arrivals to obtain display times.

\section{Image Costs and Search Validation}\label{supp:B}

\subsection{Image Sensitivities}\label{supp:B1}

Main Section 3.4 evaluates two image effects: attribute quantization error and changes in Gaussian availability. Both use sensitivities computed once from $S_0$ at fixed absolute budgets. For view $v$ at node $k$, $r_{vk}$, $D_{vk}$, and $J_{vk}$ denote the image residual, RGB MSE, and coding-attribute Jacobian. With training cameras $\mathcal V$ and node subset $\mathcal V_k$, $\omega_k=w_k|\mathcal V|/|\mathcal V_k|$ combines quadrature and camera weights. Attribute errors are evaluated with the partial-and-Full metric
\begin{equation}
\begin{aligned}
M=\Pi\Big[&(1-\mu)\sum_k\omega_k\sum_{v\in\mathcal V_k}\frac{J_{vk}^TJ_{vk}}{\max(D_{vk},10^{-12})}\\
&+\mu\sum_{v\in\mathcal V}\frac{J_{v,\mathrm{Full}}^TJ_{v,\mathrm{Full}}}{\max(D_{v,\mathrm{Full}},10^{-12})}\Big].
\end{aligned}
\tag{S6}
\end{equation}

The projection $\Pi$ retains one $7\times7$ intrinsic-coordinate block per Gaussian and diagonal entries for other attributes. Partial node $k=0,\ldots,5$ uses every sixth camera starting at $k$ in the unshuffled training list, while Full uses all cameras; one random-sign probe per assigned view estimates the Gram contribution.

The same reference renderings also evaluate changes in Gaussian inclusion. For this second effect, inclusion Jacobian $j_{vki}$ gives the completion coefficients used in Equation (7):
\begin{equation}
\begin{aligned}
g_{ki}&=\frac{|\mathcal V|}{|\mathcal V_k|}\sum_{v\in\mathcal V_k}\frac{j_{vki}^Tr_{vk}}{\max(D_{vk},10^{-12})},\\
h_{ki}&=\frac{|\mathcal V|}{|\mathcal V_k|}\sum_{v\in\mathcal V_k}\frac{\|j_{vki}\|^2}{\max(D_{vk},10^{-12})}.
\end{aligned}
\tag{S7}
\end{equation}

Equation (7) applies partial mass and quadrature weights to these coefficients once. At a cutoff where a Gaussian is not yet decoded, the encoder evaluates its inclusion derivative at zero using preceding transmittance, ungated alpha, and contrast with the color composited behind it. The estimated completion cost therefore depends on visibility and the reference partial image, rather than Gaussian count alone.

The score combines blockwise attribute-error quadratics with separate linear--quadratic inclusion changes, retaining within-Gaussian intrinsic-coordinate coupling. Cross-record and attribute--inclusion couplings, linear attribute terms, and within-prefix S+O changes are outside this approximation.

\subsection{Continuous Reception}\label{supp:B2}

For coding configuration $p$ relative to initialization $p_0$, header length $H(p)$ and record lengths $r_j(p)$ determine each cumulative completion shift:
\begin{equation}
\begin{aligned}
C_i(p)-C_i(p_0)={}&H(p)-H(p_0)\\
&+\sum_{j\le i}\big[r_j(p)-r_j(p_0)\big].
\end{aligned}
\tag{S8}
\end{equation}

With each prefix rendering held fixed, let $D_i$ be the loss after the first $i$ Gaussians and $\bar F(x)=\Pr(q\ge x)$ the probability of receiving encoded position $x$. The expected partial loss is
\begin{equation}
\mathbb E[D(q)]=D_0+\sum_i(D_i-D_{i-1})\bar F(C_i).
\tag{S9}
\end{equation}

Taking expectations of $D(q)=D_0+\sum_i(D_i-D_{i-1})\mathbf1[q\ge C_i]$ gives Equation (S9): each image transition is weighted by the probability of reaching its boundary.

For continuous requests, $\bar F^{\prime}(x)=-f_Q(x)$ and $\partial C_i/\partial r_j=\mathbf1[i\ge j]$. Differentiating Equation (S9) with fixed prefix losses and header gives main Equation (5).

\textbf{Objective evaluation.} The implemented objective combines this reception model with the image and rate terms in main Equation (4). Its image loss $\mathcal D(S,q)=\frac12\sum_v n_v\ln D_v(S,q)$ weights RGB MSE $D_v$ by RGB component count $n_v$ under a fitted-variance Gaussian likelihood. Expected delivered bits are converted to the same logarithmic scale:
\begin{equation}
\begin{aligned}
\mathcal R(L)=\ln2\Big[&(1-\mu)\int_{1/8}^{1}\min(q(b),L)p(b)\,db\\
&+\mu L\Big].
\end{aligned}\tag{S10}
\end{equation}

The initial stream fixes requests $q(b)$ for integration in log reception fraction, which weights equal proportional byte increases equally. Over $t=\ln b\in[-\ln8,0]$, two Gauss--Legendre nodes in each of three equal intervals approximate the partial integral, with Full evaluated separately. Equation (4) gives equal partial-node weights at approximately 0.145, 0.216, 0.289, 0.432, 0.579, and 0.864 of $L_0$, together with Full mass $1/2$.

The initial stream also calibrates $\kappa$ in main Equation (6); requests exceeding a candidate's length are capped at $L$. Boundary predictions accumulate all adopted length changes, including header changes. At these byte-rounded budgets, Equation (7) scores finite inclusion changes using rendering sensitivities. Doubling the quadrature from six to twelve nodes changes the measured objective by at most $0.0372\%$ of its six-node magnitude across the evaluated files.

\subsection{Allocation Search and Candidate Validation}\label{supp:B3}

Validation extends main Section 4.5 by checking completion-boundary error (Table S2), objective-change prediction (Figure A1a), and candidate selection (Figure A1b; Table S2a). Evaluation uses actual boundaries and recomputed S+O.

Tables S2 and S2a cover 56 encoded configurations from 18 search states in Bonsai, Train, and DrJohnson. Figure A1 extends the same early/middle/late precision and context search states to all 13 scenes (242 configurations, 78 pools). Each pool includes the current setting and all captured alternatives. Predictions precede encoding; measured objectives use common training views and absolute budgets.

\begin{figure*}[!t]
\captionsetup{font=small}
\centering
\includegraphics[width=\textwidth]{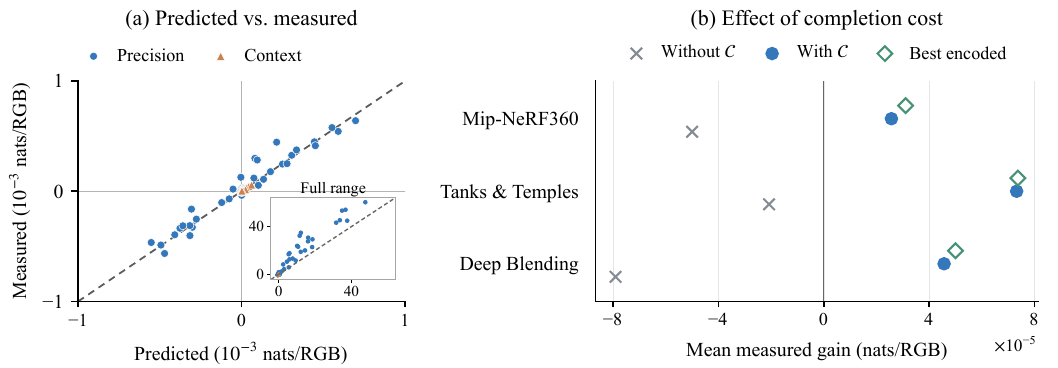}
\renewcommand{\thefigure}{A1}\caption{Candidate selection across 13 scenes. (a) Predicted versus measured changes near zero; inset shows the full range. (b) Dataset-mean gains relative to current settings; best denotes the best encoded candidate in each captured pool \cite{mipnerf360,tanks,deepblending}.}\label{fig:A1}
\end{figure*}

Figure A1(b) compares identical pools, omitting only $\mathcal C$ from Equation (7) for the control. Negative changes in (a) and positive gains in (b) indicate improvement. Values are per training RGB component; (b) averages six pools per scene and scenes equally per dataset.

\textbf{Completion-boundary accuracy.} Table S2 measures byte-position MAE within eight records on either side of each actual reception cutoff. Count MAE compares predicted and decoded completed-record counts over the same search states.

\textbf{Encoded candidate gains.} Table S2a reports the five precision pools with improving alternatives. Selected updates recover 92.6\% of their aggregate available improvement, with updates adopted in all three middle-stage pools and the current setting retained in the two late-stage pools.

\setlength{\textfloatsep}{8pt}\setlength{\dbltextfloatsep}{8pt}
\section{Allocation Analysis}\label{supp:C}

\begin{table*}[!t]
\begingroup\captionsetup{type=table,skip=6pt}
\centering\small
\setlength{\tabcolsep}{3pt}\renewcommand{\arraystretch}{1.0}
\renewcommand{\thetable}{S3}\caption{Allocation controls at equal partial bytes: A+C minus control PSNR (dB), BD-rate (\%), and control Full size (MB) \cite{mipnerf360,tanks,deepblending}.}\label{tab:S3}
\begin{tabular*}{\textwidth}{@{\extracolsep{\fill}}lrrrrrrr@{}}
\toprule
Control & 1/8 & 1/4 & 1/2 & 3/4 & Full & BD-rate & MB\\\midrule
\multicolumn{8}{l}{\textbf{Mip-NeRF360}}\\
Attribute-only, fixed weight & +2.11 & +0.67 & +0.13 & +0.04 & -0.13 & -19.84 & 8.41\\
Attribute-only, size matched & +0.85 & +0.25 & -0.01 & -0.05 & -0.05 & -8.90 & 7.28\\
Uniform completion & +0.92 & +0.27 & +0.00 & -0.04 & -0.05 & -9.25 & 7.37\\
Interval completion & +0.92 & +0.27 & +0.00 & -0.04 & -0.05 & -9.21 & 7.38\\
Order-only schedule & +1.45 & +1.07 & +0.58 & +0.40 & +0.35 & -37.26 & 7.28\\
\addlinespace[3pt]
\multicolumn{8}{l}{\textbf{Tanks \& Temples}}\\
Attribute-only, fixed weight & +1.45 & +0.37 & +0.05 & -0.01 & -0.07 & -23.02 & 8.43\\
Attribute-only, size matched & +0.48 & +0.11 & -0.02 & -0.04 & -0.04 & -8.47 & 7.20\\
Uniform completion & +0.52 & +0.12 & -0.02 & -0.04 & -0.04 & -9.07 & 7.23\\
Interval completion & +0.51 & +0.12 & -0.02 & -0.03 & -0.04 & -9.09 & 7.36\\
Order-only schedule & +0.28 & +0.16 & +0.03 & +0.00 & +0.00 & -9.73 & 7.22\\
\addlinespace[3pt]
\multicolumn{8}{l}{\textbf{Deep Blending}}\\
Attribute-only, fixed weight & +2.89 & +0.57 & +0.01 & -0.04 & -0.08 & -31.53 & 8.76\\
Attribute-only, size matched & +1.43 & +0.21 & -0.04 & -0.05 & -0.05 & -17.92 & 7.26\\
Uniform completion & +1.56 & +0.25 & -0.04 & -0.05 & -0.05 & -19.94 & 7.44\\
Interval completion & +1.56 & +0.25 & -0.04 & -0.05 & -0.05 & -19.89 & 7.43\\
Order-only schedule & +0.73 & +0.19 & +0.04 & +0.04 & +0.05 & -16.99 & 7.21\\
\bottomrule
\end{tabular*}
\endgroup
\end{table*}

\subsection{Allocation Controls}\label{supp:C1}

Main Section 4.3 tests Gaussian-specific image information against simpler allocation rules. Table S3 gives the reception-budget results for its attribute-only, averaged-completion, and order-only controls on the fixed prepared representation. The attribute-only control shares initialization, search variables, and a fixed rate weight with \mbox{$A+C$}; its size-matched variant tunes that weight and supplies the attribute-only baseline in main Table 3.

\textbf{Uniform and interval completion.} These controls retain the precision variables and search while averaging the image-derived completion costs. Uniform completion averages insertion and removal costs separately across Gaussians at each node. Interval completion averages within four equal-count order blocks, distinct from the coding intervals in A.2. The averaging preserves signed group totals and reception weights; cumulative averaged costs then score prefix changes.

\vspace{8pt}
\par\begingroup\nolinenumbers\noindent\begin{minipage}{\columnwidth}

\captionsetup{type=table}\centering\small
\setlength{\tabcolsep}{1.4pt}\renewcommand{\arraystretch}{1.04}
\renewcommand{\thetable}{S2}\caption{Mean absolute error near reception cutoffs, in completion bytes and decoded-record count \cite{mipnerf360,tanks,deepblending}.}\label{tab:S2}
\begin{tabular*}{\columnwidth}{@{\extracolsep{\fill}}lrr@{}}
\toprule
Update & Byte MAE & Count MAE\\\midrule
Precision & 123.49 & 9.27\\
Context & 25.52 & 1.81\\
\bottomrule\end{tabular*}

\end{minipage}\par\endgroup
\vspace{6pt}
\par\begingroup\nolinenumbers\noindent\begin{minipage}{\columnwidth}

\captionsetup{type=table}\centering\small
\setlength{\tabcolsep}{1.4pt}\renewcommand{\arraystretch}{1.04}
\renewcommand{\thetable}{S2a}\caption{Measured objective reductions within the encoded precision pools ($10^3$ units). Selected and best gains are relative to each current setting \cite{mipnerf360,tanks,deepblending}.}\label{tab:S2a}
\begin{tabular*}{\columnwidth}{@{\extracolsep{\fill}}llrrr@{}}
\toprule
Scene & Stage & Selected & Best & Captured (\%)\\\midrule
Bonsai & Middle & 386.47 & 498.46 & 77.53\\
Train & Middle & 950.98 & 950.98 & 100.00\\
DrJohnson & Middle & 130.60 & 130.60 & 100.00\\
Train & Late & 0.00 & 0.27 & 0.00\\
DrJohnson & Late & 0.00 & 4.53 & 0.00\\
\bottomrule\end{tabular*}

\end{minipage}\par\endgroup

\Needspace{8\baselineskip}
\textbf{Count-aware control.} Count-aware replaces image-dependent completion costs with an equal reward for each additional decoded Gaussian at an early byte budget, up to a target above the \mbox{$A+C$} count. Its rate weight is tuned to match the complete-file size, and selection uses count and size criteria.

\begin{table}[!t]
\centering\small
\setlength{\tabcolsep}{2pt}\renewcommand{\arraystretch}{1.05}
\renewcommand{\thetable}{S3a}\caption{Size-matched A+C versus Count-aware at 1/8 reception. PSNR gain is A+C minus Count-aware; SD is the sample standard deviation. BD-rate: \% \cite{mipnerf360,tanks,deepblending}.}\label{tab:S3a}
\begin{tabular*}{\linewidth}{@{\extracolsep{\fill}}lrrrr@{}}
\toprule
\multirow{2}{*}{Dataset} & \multicolumn{3}{c}{\shortstack{1/8 PSNR gain\\(dB)}} & \multirow{2}{*}{\shortstack{Mean BD-rate\\(\%)}}\\
\cmidrule(lr){2-4}
 & Mean & SD & Median & \\\midrule
Mip-NeRF360 & +0.65 & 1.02 & +0.29 & -24.03\\
Tanks \& Temples & +0.37 & 0.32 & +0.37 & -15.22\\
Deep Blending & +0.22 & 0.08 & +0.22 & -18.68\\
\bottomrule\end{tabular*}
\end{table}

Specifically, the attribute-and-rate objective includes $\alpha(1-\mu)\max(0,K-N_S(q_\star))$, where $q_\star$ is the byte-aligned \mbox{$A+C$} 1/8 budget, $N_S(q)$ counts completed Gaussians, and $K=\lceil1.05N_{A+C}(q_\star)\rceil$. For retained count $N$, $\alpha=fN_{\mathrm{RGB}}/(2N)$ with $f\in\{16,64,256\}$. Eligible streams exceed the \mbox{$A+C$} count and match Full size within 0.5\%; selection minimizes distance to $K$, then size error, then $f$. Table S3a reports the resulting quality.

\textbf{Order-only schedule.} This control assigns precision from position in the fixed order.

For interval $k$ among $K$ intervals, let $r_k=k/(K-1)$ and $m_j$ be component $j$'s median initial log-step exponent. The additional multiplier is $2^{e_{k,j}}$, where
\begin{equation}
e_{k,j}=\operatorname{clip}_{[0,9]}\!\left(\operatorname{round}\left[m_j+\beta+s(1-r_k)\right]\right).
\tag{S11}
\end{equation}

Positive $s$ makes early intervals coarser; $\beta$ sets overall precision. Search starts from slopes $0,1,2,3$ and offsets $-1,0,1,2$, then refines the offset toward the target Full size. Encoded candidates are rendered on training views and scored with Equation (4). Selection minimizes measured objective within $0.5\%$ size error, falling back to the closest size with objective as tie-breaker.

For the size-matched $A$/\mbox{$A+C$} pair, Figure A2 relates changes in allocated precision to the resulting differences in completed Gaussian counts.

\setcounter{topnumber}{1}
\begin{figure}[!hb]

\captionsetup{type=figure,skip=2pt}
\centering
\includegraphics[width=.94\columnwidth]{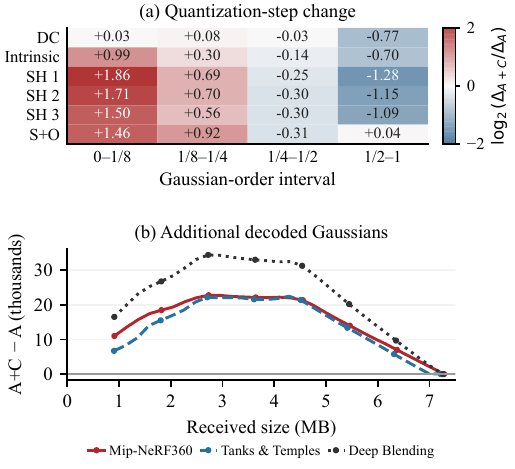}
\renewcommand{\thefigure}{A2}\caption{Size-matched A+C versus attribute-only \cite{mipnerf360,tanks,deepblending}. (a) Step changes (positive: coarser); (b) extra Gaussians at equal bytes.}\label{fig:A2}

\end{figure}

\subsection{Conditional Reconstruction Effects}\label{supp:C2}

Main Section 4.4 examines reconstruction changes by crossing quantized attributes and completion boundaries. Table S4 extends that conditional analysis across reception for the separate $A$/\mbox{$A+C$} pair optimized with the same rate weight. In PSNR $P_{x,y,s}$, $x$ supplies attributes, $y$ supplies boundaries, and $s$ selects S+O (1: on; 0: off). Subscripts $A$ and $C$ denote attribute-only and completion-aware allocation. With S+O enabled, the boundary map determines both the decoded set and its correction. The conditional contrasts are
\begin{equation}
\begin{aligned}
\Delta_{\mathrm{set},0}&=P_{C,C,0}-P_{C,A,0},\\
\Delta_{\mathrm{attr},0}&=P_{C,C,0}-P_{A,C,0},\\
\Delta_{\mathrm{S+O}}&=P_{C,C,1}-P_{C,C,0},\\
\Delta_{\mathrm{set},1}&=P_{C,C,1}-P_{C,A,1}.
\end{aligned}
\tag{S12}
\end{equation}

\begin{table}[!t]
\centering\small
\setlength{\tabcolsep}{2.2pt}\renewcommand{\arraystretch}{1.0}
\renewcommand{\thetable}{S4}\caption{Conditional PSNR contrasts (dB), Equation (S12), for $A$/\mbox{$A+C$} using the same rate weight. Set: completion-boundary map; Attr.: attribute precision. Off/on denotes S+O; the last column enables S+O with attributes and boundaries fixed.}\label{tab:S4}
\begin{tabular*}{\columnwidth}{@{\extracolsep{\fill}}lrrrr@{}}
\toprule
\multirow{2}{*}{Budget} & \multicolumn{2}{c}{Decoded set} & Attr. & S+O\\
\cmidrule(lr){2-3}
 & Off & On & Off & On $-$ Off\\\midrule
\multicolumn{5}{l}{\textbf{Mip-NeRF360} \cite{mipnerf360}}\\
1/8 & +2.44 & +2.16 & -0.05 & +2.26\\
1/4 & +1.33 & +0.79 & -0.09 & +1.22\\
1/2 & +0.63 & +0.28 & -0.12 & +0.33\\
3/4 & +0.32 & +0.18 & -0.13 & +0.04\\
Full & +0.00 & +0.00 & -0.13 & +0.00\\
\addlinespace[4pt]
\multicolumn{5}{l}{\textbf{Tanks \& Temples} \cite{tanks}}\\
1/8 & +1.89 & +1.50 & -0.03 & +1.45\\
1/4 & +0.76 & +0.43 & -0.05 & +0.57\\
1/2 & +0.24 & +0.12 & -0.07 & +0.11\\
3/4 & +0.10 & +0.06 & -0.07 & +0.01\\
Full & +0.00 & +0.00 & -0.07 & +0.00\\
\addlinespace[4pt]
\multicolumn{5}{l}{\textbf{Deep Blending} \cite{deepblending}}\\
1/8 & +3.91 & +2.90 & -0.05 & +1.28\\
1/4 & +1.33 & +0.67 & -0.08 & +0.31\\
1/2 & +0.23 & +0.11 & -0.08 & +0.01\\
3/4 & +0.07 & +0.05 & -0.08 & -0.01\\
Full & +0.00 & +0.00 & -0.08 & +0.00\\
\bottomrule\end{tabular*}

\par\vspace{10pt}
\captionsetup{type=table,skip=3pt}

\centering\small
\setlength{\tabcolsep}{2pt}\renewcommand{\arraystretch}{1.0}
\renewcommand{\thetable}{S5}\caption{Final minus $S_0$ objective ($10^6$ units) \cite{mipnerf360,tanks,deepblending}.}\label{tab:S5}
\begin{tabular*}{\columnwidth}{@{\extracolsep{\fill}}lrrrr@{}}
\toprule
Dataset & Partial & Full & Rate & Total\\\midrule
Mip-NeRF360 & -17.63 & +4.74 & +0.12 & -12.77\\
Tanks \& Temples & -13.86 & +11.64 & -0.05 & -2.27\\
Deep Blending & -21.41 & +11.91 & -0.52 & -10.01\\
\bottomrule\end{tabular*}

\end{table}

With attributes fixed, the A+C boundary map improves early PSNR in both S+O settings.

\subsection{Intermediate Reconstructions}\label{supp:C3}

Main Figure 6 compares A+C, attribute-only, Count-aware, and the order-only schedule at 29 directly decoded and rendered states: $n/32$, $n=4,\ldots,31$, and Full. Each allocation uses the same stream across cutoffs. Figure A3 shows the proposed codec's progression across all 13 scenes.

\par\medskip\noindent
\subsection{Measured Continuous Objective}\label{supp:C4}
Table S5 decomposes the measured change in Equation (4) into partial-reception, Full, and rate terms. The final allocation lowers their sum on all three datasets.

\setcounter{topnumber}{2}

\subsection{Quality and Size Across Methods}\label{supp:C5}
Table S5a reports quality and size.
\setlength{\dbltextfloatsep}{12pt}
\begin{table*}[!t]
\begingroup
\captionsetup{type=table,skip=6pt}
\centering\footnotesize
\setlength{\tabcolsep}{1.4pt}\renewcommand{\arraystretch}{1.0}
\renewcommand{\thetable}{S5a}\caption{Quality and size at evaluated operating points, including LapisGS and Matryoshka Gaussian Splatting (MGS) \cite{lapis,matryoshka}. PSNR: dB; size: MB. LapisGS levels res8--res1 use the res1 evaluation resolution; $N$ denotes the MGS Full Gaussian count.}\label{tab:S5a}
\begin{tabular*}{\textwidth}{@{\extracolsep{\fill}}llrrrrrrrrrrrr@{}}
\toprule
\multirow{2}{*}{Method} & \multirow{2}{*}{Point} & \multicolumn{4}{c}{Mip-NeRF360 \cite{mipnerf360}} & \multicolumn{4}{c}{Tanks \& Temples \cite{tanks}} & \multicolumn{4}{c}{Deep Blending \cite{deepblending}}\\
\cmidrule(lr){3-6}\cmidrule(lr){7-10}\cmidrule(lr){11-14}
 & & PSNR$\uparrow$ & SSIM$\uparrow$ & LPIPS$\downarrow$ & MB$\downarrow$ & PSNR$\uparrow$ & SSIM$\uparrow$ & LPIPS$\downarrow$ & MB$\downarrow$ & PSNR$\uparrow$ & SSIM$\uparrow$ & LPIPS$\downarrow$ & MB$\downarrow$\\\midrule
3DGS \cite{kerbl2023} & Full & 27.80 & 0.8264 & 0.2015 & 640.96 & 23.49 & 0.8308 & 0.2350 & 449.01 & 29.67 & 0.9025 & 0.2413 & 608.86\\
\addlinespace[2pt]
\multirow{5}{*}{Ours (A+C)} & 1/8 & 24.92 & 0.6603 & 0.4098 & 0.91 & 21.62 & 0.7085 & 0.4067 & 0.90 & 28.26 & 0.8617 & 0.3412 & 0.91\\
 & 1/4 & 26.54 & 0.7310 & 0.3415 & 1.82 & 23.02 & 0.7619 & 0.3480 & 1.80 & 29.28 & 0.8848 & 0.3000 & 1.81\\
 & 1/2 & 27.29 & 0.7720 & 0.2888 & 3.64 & 23.48 & 0.7898 & 0.3054 & 3.61 & 29.58 & 0.8948 & 0.2752 & 3.63\\
 & 3/4 & 27.62 & 0.7928 & 0.2577 & 5.46 & 23.60 & 0.8002 & 0.2847 & 5.41 & 29.65 & 0.8981 & 0.2626 & 5.44\\
 & Full & 27.76 & 0.8000 & 0.2446 & 7.28 & 23.64 & 0.8047 & 0.2739 & 7.21 & 29.68 & 0.8990 & 0.2567 & 7.26\\
\addlinespace[2pt]
\multirow{8}{*}{GoDe \cite{gode}} & L0 & 25.50 & 0.7280 & 0.3401 & 4.49 & 21.57 & 0.7518 & 0.3524 & 4.08 & 28.88 & 0.8800 & 0.3050 & 4.40\\
 & L1 & 26.31 & 0.7597 & 0.3048 & 6.46 & 22.29 & 0.7695 & 0.3294 & 5.50 & 29.27 & 0.8870 & 0.2903 & 6.03\\
 & L2 & 26.89 & 0.7837 & 0.2743 & 9.44 & 22.79 & 0.7835 & 0.3097 & 7.45 & 29.43 & 0.8907 & 0.2797 & 8.28\\
 & L3 & 27.28 & 0.8018 & 0.2480 & 14.00 & 23.16 & 0.7955 & 0.2919 & 10.17 & 29.51 & 0.8929 & 0.2715 & 11.44\\
 & L4 & 27.54 & 0.8144 & 0.2272 & 20.92 & 23.40 & 0.8046 & 0.2774 & 14.00 & 29.53 & 0.8939 & 0.2653 & 15.84\\
 & L5 & 27.69 & 0.8221 & 0.2120 & 31.38 & 23.55 & 0.8109 & 0.2658 & 19.35 & 29.54 & 0.8944 & 0.2607 & 21.94\\
 & L6 & 27.76 & 0.8261 & 0.2022 & 46.61 & 23.60 & 0.8143 & 0.2585 & 26.66 & 29.54 & 0.8944 & 0.2577 & 30.27\\
 & L7 & 27.76 & 0.8274 & 0.1977 & 68.27 & 23.62 & 0.8165 & 0.2537 & 36.96 & 29.53 & 0.8942 & 0.2563 & 41.51\\
\addlinespace[2pt]
\multirow{4}{*}{\shortstack[l]{PCGS \cite{pcgs}\\High-rate}} & ss0 & 27.94 & 0.8209 & 0.2171 & 13.51 & 23.84 & 0.8219 & 0.2576 & 6.88 & 29.99 & 0.9024 & 0.2653 & 3.83\\
 & ss1 & 28.05 & 0.8231 & 0.2147 & 17.84 & 23.92 & 0.8254 & 0.2535 & 9.08 & 30.15 & 0.9052 & 0.2613 & 5.13\\
 & ss2 & 28.07 & 0.8236 & 0.2141 & 22.71 & 23.94 & 0.8260 & 0.2527 & 11.93 & 30.18 & 0.9057 & 0.2604 & 6.85\\
 & ss3 & 28.07 & 0.8237 & 0.2137 & 28.04 & \textemdash{} & \textemdash{} & \textemdash{} & \textemdash{} & \textemdash{} & \textemdash{} & \textemdash{} & \textemdash{}\\
\addlinespace[2pt]
\multirow{4}{*}{\shortstack[l]{PCGS \cite{pcgs}\\Low-rate}} & ss0 & 25.33 & 0.7201 & 0.3613 & 1.93 & 22.74 & 0.7807 & 0.3199 & 2.52 & 28.53 & 0.8816 & 0.3109 & 1.48\\
 & ss1 & 25.95 & 0.7388 & 0.3473 & 2.46 & 23.10 & 0.7929 & 0.3067 & 3.23 & 29.19 & 0.8886 & 0.3032 & 1.76\\
 & ss2 & 26.18 & 0.7488 & 0.3371 & 3.33 & 23.22 & 0.7980 & 0.2998 & 4.80 & 29.43 & 0.8921 & 0.2969 & 2.46\\
 & ss3 & 26.38 & 0.7610 & 0.3211 & 4.97 & \textemdash{} & \textemdash{} & \textemdash{} & \textemdash{} & \textemdash{} & \textemdash{} & \textemdash{} & \textemdash{}\\
\addlinespace[2pt]
\multirow{5}{*}{ProGS \cite{progs}} & lv0 & 24.65 & 0.6855 & 0.3652 & 8.92 & 18.64 & 0.6545 & 0.4866 & 3.56 & 27.19 & 0.8517 & 0.3634 & 1.57\\
 & lv1 & 26.44 & 0.7625 & 0.2695 & 14.89 & 20.75 & 0.7112 & 0.4157 & 6.08 & 28.91 & 0.8834 & 0.3006 & 2.56\\
 & lv2 & 27.22 & 0.8045 & 0.2223 & 18.34 & 22.29 & 0.7663 & 0.3348 & 8.21 & 29.24 & 0.8931 & 0.2813 & 3.55\\
 & lv3 & 27.54 & 0.8233 & 0.2030 & 21.16 & 23.35 & 0.8129 & 0.2618 & 10.56 & 29.18 & 0.8930 & 0.2807 & 5.04\\
 & lv4 & 27.56 & 0.8244 & 0.2018 & 22.52 & 23.81 & 0.8289 & 0.2387 & 12.20 & 29.18 & 0.8930 & 0.2807 & 6.46\\
\addlinespace[2pt]
\multirow{6}{*}{RDO-Gaussian \cite{rdo}} & rate5 & 24.53 & 0.6920 & 0.3933 & 1.86 & 21.77 & 0.7410 & 0.3790 & 1.39 & 28.35 & 0.8714 & 0.3326 & 1.30\\
 & rate4 & 25.42 & 0.7424 & 0.3300 & 3.77 & 22.41 & 0.7745 & 0.3326 & 2.56 & 29.11 & 0.8879 & 0.2978 & 2.54\\
 & rate3 & 26.16 & 0.7733 & 0.2856 & 6.40 & 22.59 & 0.7914 & 0.3051 & 4.11 & 29.37 & 0.8952 & 0.2780 & 4.35\\
 & rate2 & 26.66 & 0.7939 & 0.2532 & 10.03 & 22.78 & 0.8023 & 0.2852 & 6.19 & 29.50 & 0.8994 & 0.2651 & 7.03\\
 & rate1 & 27.04 & 0.8075 & 0.2313 & 15.43 & 22.99 & 0.8101 & 0.2712 & 9.38 & 29.49 & 0.9011 & 0.2572 & 11.58\\
 & rate0 & 27.26 & 0.8134 & 0.2217 & 23.68 & 23.00 & 0.8135 & 0.2644 & 14.06 & 29.59 & 0.9020 & 0.2522 & 18.92\\
\addlinespace[2pt]
\multirow{4}{*}{LapisGS \cite{lapis}} & L0 & 17.66 & 0.4123 & 0.5364 & 252.28 & 17.09 & 0.5937 & 0.4618 & 125.48 & 23.04 & 0.7348 & 0.4126 & 146.39\\
 & L1 & 21.31 & 0.5567 & 0.4335 & 606.46 & 19.81 & 0.6868 & 0.3836 & 281.66 & 26.25 & 0.8191 & 0.3214 & 277.64\\
 & L2 & 24.87 & 0.6996 & 0.3182 & 1069.62 & 22.22 & 0.7633 & 0.3166 & 513.50 & 28.36 & 0.8725 & 0.2517 & 432.54\\
 & L3 & 26.54 & 0.7548 & 0.2702 & 1486.03 & 23.16 & 0.7911 & 0.2878 & 794.66 & 29.05 & 0.8900 & 0.2272 & 588.57\\
\addlinespace[2pt]
\multirow{4}{*}{MGS \cite{matryoshka}} & 0.05N & 24.24 & 0.6531 & 0.3294 & 59.00 & 21.42 & 0.7371 & 0.3162 & 59.00 & 27.86 & 0.8788 & 0.2195 & 59.00\\
 & 0.10N & 25.84 & 0.7252 & 0.2376 & 118.00 & 22.70 & 0.7873 & 0.2362 & 118.00 & 28.55 & 0.8945 & 0.1883 & 118.00\\
 & 0.20N & 27.20 & 0.7931 & 0.1677 & 236.00 & 23.52 & 0.8242 & 0.1803 & 236.00 & 28.90 & 0.9040 & 0.1703 & 236.00\\
 & Full & 28.27 & 0.8416 & 0.1285 & 1180.00 & 24.29 & 0.8556 & 0.1352 & 1180.00 & 29.01 & 0.9069 & 0.1657 & 1180.00\\
\bottomrule
\end{tabular*}
\endgroup
\end{table*}

\FloatBarrier
\Needspace{8\baselineskip}
\subsection{S+O and Completion-Aware Allocation}\label{supp:C6}
\label{sec:factorial}
Main Section 4.4 analyzes S+O through crossed reconstructions. To examine whether completion-aware allocation also remains effective when preparation omits correction, Table S11 compares $A$ and \mbox{$A+C$} with S+O enabled or disabled throughout preparation and decoding. Both preparations start from the same 15K checkpoints and follow the same 25K joint and 5K final refinement schedule. Within each S+O setting, the allocators share the prepared representation, initial coding settings, reference sensitivities, and search domain. All four variants use the same rate weight and absolute optimization budgets.

Without S+O, \mbox{$A+C$} improves 1/8 PSNR over $A$ by 2.75, 1.90, and 3.30 dB on the three datasets, with gains in all 13 scenes. The corresponding BD-rate reductions are 22.00\%, 27.45\%, and 36.29\% (Table S12). Combining S+O with \mbox{$A+C$} further reduces BD-rate by 16.19\%, 13.19\%, and 9.61\% relative to \mbox{$A+C$} without S+O. Thus, completion-aware allocation remains effective without correction, while the combined codec provides additional rate--distortion gains.

\onecolumn
\setlength{\dbltextfloatsep}{20pt}
\section{Per-Scene Reconstruction Results}\label{supp:D}
\begingroup\captionsetup{type=figure}\centering
\includegraphics[page=1,width=\textwidth,keepaspectratio]{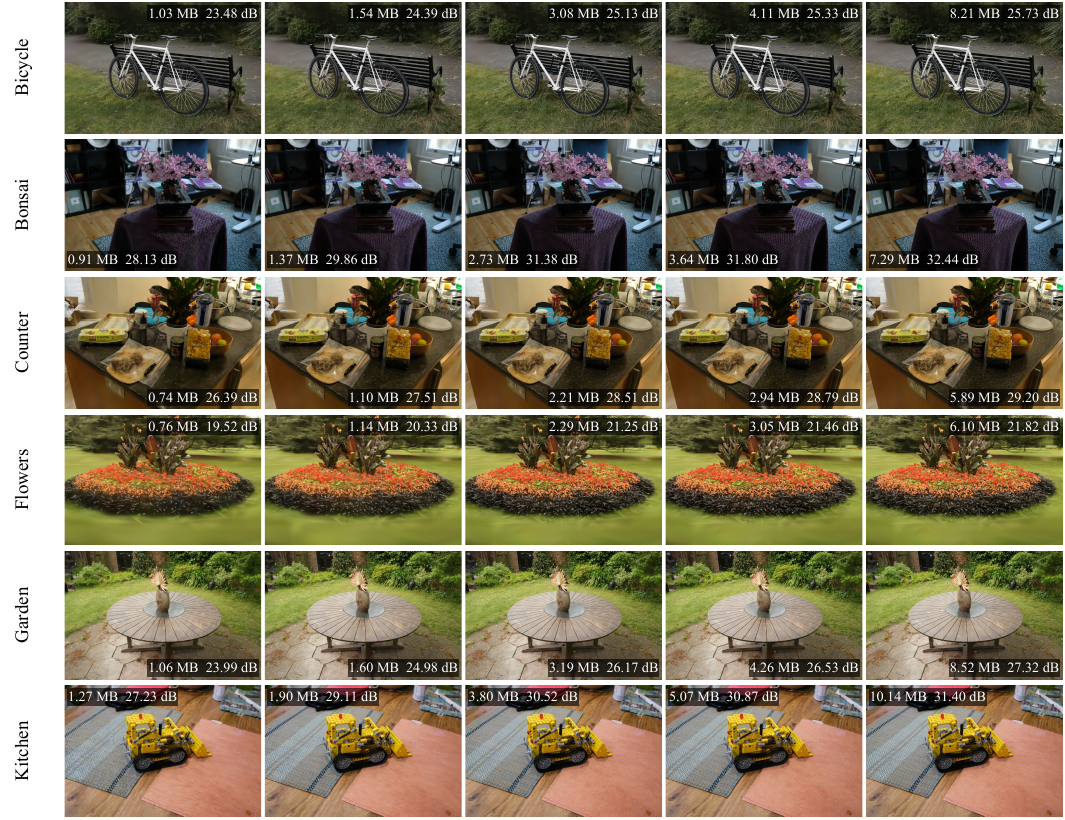}
\renewcommand{\thefigure}{A3}\caption{Progressive reconstruction across all 13 scenes \cite{mipnerf360,tanks,deepblending}. Each row uses the same scene, camera, and stream across reception budgets.}\label{fig:A3}

\par\endgroup
\smallskip
\par\begingroup\captionsetup{type=table,skip=5pt}\centering\renewcommand{\thetable}{S6}\caption{Per-scene reconstruction from the final Mip-NeRF360 streams \cite{mipnerf360}. PSNR is in dB and size in MB.}\label{tab:S6}\par\endgroup
\noindent
\begin{minipage}[t]{.48\textwidth}
\captionsetup{type=table,skip=5pt}\centering\small
\setlength{\tabcolsep}{1pt}\renewcommand{\arraystretch}{1.10}
\begin{tabular*}{\linewidth}{@{\extracolsep{\fill}}llrrrrr@{}}\toprule
Scene & Metric & 1/8 & 1/4 & 1/2 & 3/4 & Full\\\midrule
\multirow{4}{*}{Bicycle} & MB & 1.03 & 2.05 & 4.11 & 6.16 & 8.21\\
 & PSNR & 23.48 & 24.82 & 25.33 & 25.62 & 25.73\\
 & SSIM & 0.5518 & 0.6483 & 0.6972 & 0.7285 & 0.7398\\
 & LPIPS & 0.4619 & 0.3817 & 0.3316 & 0.2926 & 0.2743\\
\addlinespace[2pt]
\multirow{4}{*}{Bonsai} & MB & 0.91 & 1.82 & 3.64 & 5.46 & 7.29\\
 & PSNR & 28.13 & 30.59 & 31.80 & 32.24 & 32.44\\
 & SSIM & 0.8419 & 0.8967 & 0.9227 & 0.9307 & 0.9338\\
 & LPIPS & 0.3248 & 0.2549 & 0.2128 & 0.1977 & 0.1908\\
\addlinespace[2pt]
\multirow{4}{*}{Counter} & MB & 0.74 & 1.47 & 2.94 & 4.41 & 5.89\\
 & PSNR & 26.39 & 27.99 & 28.79 & 29.08 & 29.20\\
 & SSIM & 0.7978 & 0.8463 & 0.8765 & 0.8886 & 0.8931\\
 & LPIPS & 0.3403 & 0.2812 & 0.2353 & 0.2140 & 0.2046\\

\bottomrule\end{tabular*}
\end{minipage}\hfill
\begin{minipage}[t]{.48\textwidth}
\captionsetup{type=table,skip=5pt}\centering\small
\setlength{\tabcolsep}{1pt}\renewcommand{\arraystretch}{1.10}
\begin{tabular*}{\linewidth}{@{\extracolsep{\fill}}llrrrrr@{}}\toprule
Scene & Metric & 1/8 & 1/4 & 1/2 & 3/4 & Full\\\midrule
\multirow{4}{*}{Flowers} & MB & 0.76 & 1.53 & 3.05 & 4.58 & 6.10\\
 & PSNR & 19.52 & 20.86 & 21.46 & 21.73 & 21.82\\
 & SSIM & 0.4049 & 0.4898 & 0.5436 & 0.5697 & 0.5788\\
 & LPIPS & 0.5395 & 0.4732 & 0.4238 & 0.3955 & 0.3823\\
\addlinespace[2pt]
\multirow{4}{*}{Garden} & MB & 1.06 & 2.13 & 4.26 & 6.39 & 8.52\\
 & PSNR & 23.99 & 25.56 & 26.53 & 27.07 & 27.32\\
 & SSIM & 0.5941 & 0.7040 & 0.7775 & 0.8157 & 0.8305\\
 & LPIPS & 0.4277 & 0.3382 & 0.2521 & 0.1936 & 0.1692\\
\bottomrule\end{tabular*}
\end{minipage}

\clearpage
\begingroup\captionsetup{type=figure}\centering
\includegraphics[page=2,width=\textwidth,keepaspectratio]{figures/all_scene_progressive.pdf}
\par\smallskip{\small\textbf{Figure A3 (continued).}}

\par\endgroup
\smallskip
\par\begingroup\captionsetup{type=table,skip=5pt}\centering\caption*{\textbf{Table S6 (continued).}}\par\endgroup
\noindent
\begin{minipage}[t]{.48\textwidth}
\captionsetup{type=table,skip=5pt}\centering\small
\setlength{\tabcolsep}{1pt}\renewcommand{\arraystretch}{1.10}
\begin{tabular*}{\linewidth}{@{\extracolsep{\fill}}llrrrrr@{}}\toprule
Scene & Metric & 1/8 & 1/4 & 1/2 & 3/4 & Full\\\midrule
\multirow{4}{*}{Kitchen} & MB & 1.27 & 2.53 & 5.07 & 7.60 & 10.14\\
 & PSNR & 27.23 & 29.81 & 30.87 & 31.24 & 31.40\\
 & SSIM & 0.8186 & 0.8812 & 0.9082 & 0.9173 & 0.9209\\
 & LPIPS & 0.2809 & 0.2032 & 0.1527 & 0.1351 & 0.1287\\
\addlinespace[2pt]
\multirow{4}{*}{Room} & MB & 0.78 & 1.56 & 3.12 & 4.68 & 6.24\\
 & PSNR & 29.56 & 31.02 & 31.71 & 31.99 & 32.12\\
 & SSIM & 0.8586 & 0.8878 & 0.9065 & 0.9142 & 0.9169\\
 & LPIPS & 0.3164 & 0.2720 & 0.2338 & 0.2150 & 0.2082\\

\bottomrule\end{tabular*}
\end{minipage}\hfill
\begin{minipage}[t]{.48\textwidth}
\captionsetup{type=table,skip=5pt}\centering\small
\setlength{\tabcolsep}{1pt}\renewcommand{\arraystretch}{1.10}
\begin{tabular*}{\linewidth}{@{\extracolsep{\fill}}llrrrrr@{}}\toprule
Scene & Metric & 1/8 & 1/4 & 1/2 & 3/4 & Full\\\midrule
\multirow{4}{*}{Stump} & MB & 0.89 & 1.79 & 3.57 & 5.36 & 7.14\\
 & PSNR & 24.12 & 25.67 & 26.38 & 26.73 & 26.81\\
 & SSIM & 0.5897 & 0.6821 & 0.7334 & 0.7580 & 0.7635\\
 & LPIPS & 0.4437 & 0.3660 & 0.3045 & 0.2675 & 0.2557\\
\addlinespace[2pt]
\multirow{4}{*}{Treehill} & MB & 0.75 & 1.50 & 3.00 & 4.50 & 6.00\\
 & PSNR & 21.89 & 22.51 & 22.74 & 22.90 & 22.97\\
 & SSIM & 0.4858 & 0.5427 & 0.5822 & 0.6124 & 0.6227\\
 & LPIPS & 0.5534 & 0.5029 & 0.4525 & 0.4085 & 0.3876\\
\bottomrule\end{tabular*}
\end{minipage}

\twocolumn
\begin{table}[!t]

\captionsetup{type=table}\centering\footnotesize
\setlength{\tabcolsep}{1pt}\renewcommand{\arraystretch}{1.08}
\renewcommand{\thetable}{S7}\caption{Per-scene reconstruction: Tanks \& Temples \cite{tanks} and Deep Blending \cite{deepblending}.}\label{tab:S7}
\begin{tabular*}{\linewidth}{@{\extracolsep{\fill}}llrrrrr@{}}
\toprule
Scene & Metric & 1/8 & 1/4 & 1/2 & 3/4 & Full\\\midrule
\multirow{4}{*}{Train} & MB & 0.78 & 1.56 & 3.13 & 4.69 & 6.26\\
 & PSNR & 20.59 & 21.63 & 22.01 & 22.09 & 22.12\\
 & SSIM & 0.6819 & 0.7294 & 0.7574 & 0.7678 & 0.7720\\
 & LPIPS & 0.4148 & 0.3630 & 0.3235 & 0.3050 & 0.2955\\
\addlinespace[2pt]
\multirow{4}{*}{Truck} & MB & 1.02 & 2.04 & 4.09 & 6.13 & 8.17\\
 & PSNR & 22.66 & 24.41 & 24.94 & 25.10 & 25.16\\
 & SSIM & 0.7350 & 0.7944 & 0.8223 & 0.8326 & 0.8374\\
 & LPIPS & 0.3986 & 0.3331 & 0.2874 & 0.2644 & 0.2523\\
\addlinespace[2pt]
\multirow{4}{*}{DrJohnson} & MB & 1.03 & 2.06 & 4.12 & 6.19 & 8.25\\
 & PSNR & 28.09 & 28.96 & 29.22 & 29.30 & 29.33\\
 & SSIM & 0.8567 & 0.8820 & 0.8936 & 0.8977 & 0.8988\\
 & LPIPS & 0.3429 & 0.2999 & 0.2738 & 0.2603 & 0.2542\\
\addlinespace[2pt]
\multirow{4}{*}{Playroom} & MB & 0.78 & 1.57 & 3.13 & 4.70 & 6.27\\
 & PSNR & 28.43 & 29.61 & 29.94 & 30.01 & 30.03\\
 & SSIM & 0.8666 & 0.8877 & 0.8961 & 0.8985 & 0.8991\\
 & LPIPS & 0.3395 & 0.3001 & 0.2766 & 0.2649 & 0.2592\\
\bottomrule\end{tabular*}

\end{table}
\setlength{\textfloatsep}{10pt}
\section{Comparative Decoding and Display Latency}\label{supp:E}

\textbf{Reception session.} Main Figure 4 evaluates when reconstructions become visible during transmission. Each session starts with an empty receiver. The proposed codec consumes one stream; GoDe, PCGS, and ProGS reuse received bytes, with ProGS rebuilding cumulative prefixes. The evaluated RDO-Gaussian session sends independent files in increasing-size order, so their bytes accumulate. A request for one operating point transfers only that file.

\textbf{Processing measurements.} Processing times in Table S9 are measured on an AMD EPYC 9124 CPU and an NVIDIA RTX 6000 Ada GPU, with four physical CPU cores allocated to each method. Timing starts after CUDA, renderer, and camera initialization and includes scene-specific setup, file reads, decoding, reconstruction, GPU upload, S+O, rendering, and synchronization. After one warm-up, scene medians over three cached-input runs are averaged by dataset.

\begin{nolinenumbers}
\par\Needspace{14\baselineskip}\begingroup
\captionsetup{type=table,skip=6pt}

\centering\footnotesize
\setlength{\tabcolsep}{.8pt}\renewcommand{\arraystretch}{1.10}
\renewcommand{\thetable}{S9}\caption{Processing-only time to the first and final evaluated displays (s).}\label{tab:S9}
\begin{tabular*}{\columnwidth}{@{\extracolsep{\fill}}lrrrrrr@{}}
\toprule
& \multicolumn{2}{c}{\shortstack{Mip-NeRF360\\\cite{mipnerf360}}} & \multicolumn{2}{c}{\shortstack{Tanks \&\\Temples \cite{tanks}}} & \multicolumn{2}{c}{\shortstack{Deep\\Blending \cite{deepblending}}}\\
\cmidrule(lr){2-3}\cmidrule(lr){4-5}\cmidrule(lr){6-7}
Method & First & Final & First & Final & First & Final\\\midrule
Ours (\mbox{$A+C$}) & 0.20 & 1.42 & 0.18 & 1.39 & 0.20 & 1.37\\
GoDe \cite{gode} & 0.03 & 0.58 & 0.04 & 0.34 & 0.02 & 0.30\\
PCGS (High-rate) \cite{pcgs} & 20.06 & 28.28 & 10.25 & 13.52 & 5.08 & 7.07\\
PCGS (Low-rate) \cite{pcgs} & 2.93 & 5.97 & 4.46 & 6.94 & 1.92 & 3.16\\
ProGS \cite{progs} & 8.52 & 100.12 & 3.72 & 55.23 & 1.63 & 23.39\\
RDO-Gaussian \cite{rdo} & 0.35 & 13.71 & 0.24 & 7.94 & 0.22 & 10.04\\
\bottomrule
\end{tabular*}
\par\endgroup
\end{nolinenumbers}

\newpage
\textbf{Arrival and processing.} The display-time model combines measured decoding, reconstruction, and rendering services with byte arrivals at 1, 10, and 50 Mbps. Transfer and processing overlap; an updated image becomes visible when its required data and processing are complete.

At throughput $B$ Mbps, cumulative bytes $V_j$ arrive at $a_j=8V_j/(10^6B)$. Measured entropy, reconstruction, and display services $d_j,u_j,v_j$ complete at $D_j,U_j,T_j$, respectively. Each stage waits for its input and the previous update:
\begin{equation}
\begin{aligned}
D_j&=\max(a_j,D_{j-1})+d_j,\\
U_j&=\max(D_j,U_{j-1})+u_j,\\
T_j&=\max(U_j,T_{j-1})+v_j.
\end{aligned}
\tag{S13}
\end{equation}

Header setup precedes $D_1$. Complete layers or files use the measured decode-and-display service $s_j$, with $T_j=\max(a_j,T_{j-1})+s_j$.

\textbf{TCP validation.} Quality targets are dataset-mean PSNR within 3 or 1 dB of the original 30K 3DGS reference; their times are the first qualifying evaluated displays. Bandwidth-paced TCP sessions check the display-time model at 1, 10, and 50 Mbps using these targets.

Across 70 dataset-mean quality-target times at 10 and 50 Mbps, TCP and model times differ by 1.26\% in mean absolute relative error and 7.90\% at maximum; the maximum absolute difference is 0.35 s.

\textbf{Time to comparable quality.} Figure A4 extends the display-time curves in main Figure 4 across datasets and bandwidths. Table S10 reports the first evaluated displays meeting the same quality targets.

\par\medskip\noindent
\textbf{Uncompressed 3DGS reference.} For a PLY of $s$ MB at $B$ Mbps, $T_{\mathrm{3DGS}}=8s/B+t_{\mathrm{load}}+t_{\mathrm{render}}$. Loading includes GPU upload and follows complete-file arrival; $1\,\mathrm{MB}=10^6$ bytes. Both targets use this Full display.

Native loading and rendering services average 2.83, 2.35, and 2.97 s on Mip-NeRF360, Tanks \& Temples, and Deep Blending, respectively.

\begin{figure*}[!t]
\captionsetup{type=figure,skip=6pt}

\centering
\includegraphics[width=\textwidth]{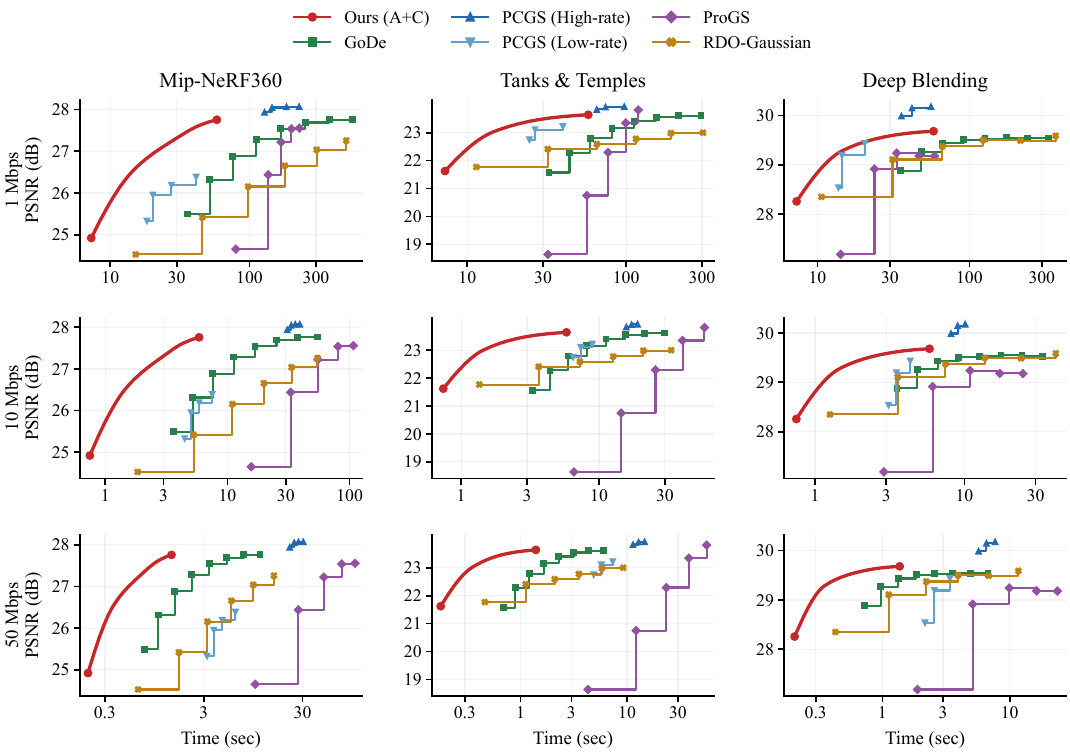}
\renewcommand{\thefigure}{A4}\caption{Quality during delivery at 1, 10, and 50 Mbps on the three datasets \cite{mipnerf360,tanks,deepblending}, with comparison codecs \cite{gode,pcgs,progs,rdo}. Stream curves interpolate evaluated display states; steps show complete level updates.}\label{fig:A4}

\end{figure*}

\begin{nolinenumbers}\par\Needspace{14\baselineskip}\begingroup\captionsetup{type=table,skip=4pt}\centering\footnotesize
\setlength{\tabcolsep}{.5pt}\renewcommand{\arraystretch}{1.06}
\captionsetup{type=table,skip=6pt}
\renewcommand{\thetable}{S10}\caption{First-display time (s) within 3 or 1 dB of the dataset-average 3DGS reference, using the criterion of Figure 4. N/A: target outside the evaluated quality range. The 3DGS rows include complete PLY transfer, loading, GPU upload, and rendering.}\label{tab:S10}
\begin{tabular*}{\columnwidth}{@{\extracolsep{\fill}}lrrrrrr@{}}
\toprule
& \multicolumn{3}{c}{Within 3 dB of 3DGS} & \multicolumn{3}{c}{Within 1 dB of 3DGS}\\
\cmidrule(lr){2-4}\cmidrule(lr){5-7}
Method & 1 Mbps & 10 Mbps & 50 Mbps & 1 Mbps & 10 Mbps & 50 Mbps\\\midrule
\multicolumn{7}{l}{\textbf{Mip-NeRF360 \cite{mipnerf360}}}\\
3DGS (full PLY) \cite{kerbl2023} & 5130.5 & 515.6 & 105.4 & 5130.5 & 515.6 & 105.4\\
Ours (\mbox{$A+C$}) & 7.3 & 0.7 & 0.2 & 29.1 & 2.9 & 0.7\\
GoDe \cite{gode} & 35.9 & 3.6 & 0.7 & 75.6 & 7.6 & 1.5\\
PCGS (High-rate) \cite{pcgs} & 128.1 & 30.9 & 22.2 & 128.1 & 30.9 & 22.2\\
PCGS (Low-rate) \cite{pcgs} & 18.3 & 4.5 & 3.2 & N/A & N/A & N/A\\
ProGS \cite{progs} & 136.4 & 32.9 & 27.2 & 168.9 & 55.0 & 49.3\\
RDO-Gaussian \cite{rdo} & 45.8 & 5.3 & 1.7 & 303.4 & 33.5 & 9.5\\

\bottomrule\end{tabular*}\par\endgroup\end{nolinenumbers}

\FloatBarrier
\begin{nolinenumbers}\begingroup\captionsetup{type=table,skip=4pt}\centering\footnotesize
\setlength{\tabcolsep}{.5pt}\renewcommand{\arraystretch}{1.06}
\caption*{\textbf{Table S10 (continued).}}
\begin{tabular*}{\columnwidth}{@{\extracolsep{\fill}}lrrrrrr@{}}
\toprule
& \multicolumn{3}{c}{Within 3 dB of 3DGS} & \multicolumn{3}{c}{Within 1 dB of 3DGS}\\
\cmidrule(lr){2-4}\cmidrule(lr){5-7}
Method & 1 Mbps & 10 Mbps & 50 Mbps & 1 Mbps & 10 Mbps & 50 Mbps\\\midrule
\multicolumn{7}{l}{\textbf{Tanks \& Temples \cite{tanks}}}\\
3DGS (full PLY) \cite{kerbl2023} & 3594.5 & 361.6 & 74.2 & 3594.5 & 361.6 & 74.2\\
Ours (\mbox{$A+C$}) & 7.2 & 0.7 & 0.2 & 14.4 & 1.5 & 0.4\\
GoDe \cite{gode} & 32.7 & 3.3 & 0.7 & 59.6 & 6.0 & 1.2\\
PCGS (High-rate) \cite{pcgs} & 65.3 & 15.8 & 11.4 & 65.3 & 15.8 & 11.4\\
PCGS (Low-rate) \cite{pcgs} & 24.6 & 6.5 & 4.9 & 24.6 & 6.5 & 4.9\\
ProGS \cite{progs} & 56.6 & 14.5 & 12.2 & 99.5 & 40.3 & 38.0\\
RDO-Gaussian \cite{rdo} & 11.4 & 1.4 & 0.5 & 65.3 & 7.3 & 2.1\\
\bottomrule\end{tabular*}\par\endgroup\end{nolinenumbers}

\par

\newpage
\begin{nolinenumbers}
\begingroup\captionsetup{type=table,skip=4pt}\centering\footnotesize
\setlength{\tabcolsep}{.5pt}\renewcommand{\arraystretch}{1.06}
\caption*{\textbf{Table S10 (continued).}}
\begin{tabular*}{\columnwidth}{@{\extracolsep{\fill}}lrrrrrr@{}}
\toprule
& \multicolumn{3}{c}{Within 3 dB of 3DGS} & \multicolumn{3}{c}{Within 1 dB of 3DGS}\\
\cmidrule(lr){2-4}\cmidrule(lr){5-7}
Method & 1 Mbps & 10 Mbps & 50 Mbps & 1 Mbps & 10 Mbps & 50 Mbps\\\midrule

\multicolumn{7}{l}{\textbf{Deep Blending \cite{deepblending}}}\\
3DGS (full PLY) \cite{kerbl2023} & 4873.9 & 490.1 & 100.4 & 4873.9 & 490.1 & 100.4\\
Ours (\mbox{$A+C$}) & 7.3 & 0.7 & 0.2 & 14.5 & 1.5 & 0.4\\
GoDe \cite{gode} & 35.2 & 3.5 & 0.7 & 35.2 & 3.5 & 0.7\\
PCGS (High-rate) \cite{pcgs} & 35.7 & 8.1 & 5.7 & 35.7 & 8.1 & 5.7\\
PCGS (Low-rate) \cite{pcgs} & 13.7 & 3.1 & 2.2 & 14.5 & 3.5 & 2.6\\
ProGS \cite{progs} & 14.2 & 2.9 & 1.9 & 23.7 & 6.1 & 5.1\\
RDO-Gaussian \cite{rdo} & 10.6 & 1.3 & 0.4 & 31.2 & 3.6 & 1.1\\
\bottomrule\end{tabular*}\par\endgroup
\end{nolinenumbers}

\clearpage
\twocolumn[{
\noindent\textbf{Additional tables for Section C.6.}\par\smallskip
\begingroup
\captionsetup{type=table,skip=6pt}
\centering\fontsize{9.5}{11.4}\selectfont
\setlength{\tabcolsep}{1pt}\renewcommand{\arraystretch}{1.62}
\renewcommand{\thetable}{S11}\caption{S+O and allocation at identical partial-reception bytes, defined by fractions of the final S+O-on \mbox{$A+C$} stream. $A$: attribute-only; \mbox{$A+C$}: completion-aware. Values are equal-weight scene means; PSNR: dB; received size: MB ($10^6$ bytes). Full file sizes depend on S+O and allocation optimization.}\label{tab:S11}
\begin{tabular*}{\textwidth}{@{\extracolsep{\fill}}llrrrrrrrrrrrr@{}}
\toprule
\multirow{2}{*}{\shortstack{S+O /\\Alloc.}} & \multirow{2}{*}{Recv.} & \multicolumn{4}{c}{Mip-NeRF360 \cite{mipnerf360}} & \multicolumn{4}{c}{Tanks \& Temples \cite{tanks}} & \multicolumn{4}{c}{Deep Blending \cite{deepblending}}\\
\cmidrule(lr){3-6}\cmidrule(lr){7-10}\cmidrule(lr){11-14}
 & & PSNR$\uparrow$ & SSIM$\uparrow$ & LPIPS$\downarrow$ & MB$\downarrow$ & PSNR$\uparrow$ & SSIM$\uparrow$ & LPIPS$\downarrow$ & MB$\downarrow$ & PSNR$\uparrow$ & SSIM$\uparrow$ & LPIPS$\downarrow$ & MB$\downarrow$\\\midrule
\multirow{5}{*}{\shortstack[l]{S+O off\\$A$}} & 1/8 & 21.43 & 0.5889 & 0.4624 & 0.91 & 19.64 & 0.6619 & 0.4452 & 0.90 & 24.39 & 0.8072 & 0.4087 & 0.91\\
 & 1/4 & 25.47 & 0.6998 & 0.3736 & 1.82 & 22.34 & 0.7381 & 0.3719 & 1.80 & 28.40 & 0.8649 & 0.3317 & 1.81\\
 & 1/2 & 26.99 & 0.7621 & 0.3044 & 3.64 & 23.26 & 0.7807 & 0.3178 & 3.61 & 29.53 & 0.8903 & 0.2852 & 3.63\\
 & 3/4 & 27.42 & 0.7868 & 0.2684 & 5.46 & 23.43 & 0.7947 & 0.2933 & 5.41 & 29.61 & 0.8956 & 0.2697 & 5.44\\
 & Full & 27.57 & 0.7976 & 0.2468 & 7.97 & 23.46 & 0.8021 & 0.2751 & 7.98 & 29.55 & 0.8972 & 0.2578 & 8.39\\
\addlinespace[4pt]
\multirow{5}{*}{\shortstack[l]{S+O off\\\mbox{$A+C$}}} & 1/8 & 24.18 & 0.6485 & 0.4167 & 0.91 & 21.53 & 0.7107 & 0.4009 & 0.90 & 27.68 & 0.8525 & 0.3503 & 0.91\\
 & 1/4 & 26.29 & 0.7281 & 0.3439 & 1.82 & 22.84 & 0.7610 & 0.3464 & 1.80 & 29.24 & 0.8827 & 0.3012 & 1.81\\
 & 1/2 & 27.11 & 0.7703 & 0.2912 & 3.64 & 23.29 & 0.7881 & 0.3060 & 3.61 & 29.51 & 0.8936 & 0.2758 & 3.63\\
 & 3/4 & 27.40 & 0.7908 & 0.2599 & 5.46 & 23.39 & 0.7979 & 0.2861 & 5.41 & 29.51 & 0.8964 & 0.2637 & 5.44\\
 & Full & 27.43 & 0.7948 & 0.2504 & 6.96 & 23.38 & 0.8007 & 0.2782 & 6.85 & 29.47 & 0.8965 & 0.2594 & 7.06\\
\addlinespace[4pt]
\multirow{5}{*}{\shortstack[l]{S+O on\\$A$}} & 1/8 & 22.81 & 0.6151 & 0.4498 & 0.91 & 20.17 & 0.6767 & 0.4359 & 0.90 & 25.37 & 0.8221 & 0.3971 & 0.91\\
 & 1/4 & 25.87 & 0.7072 & 0.3689 & 1.82 & 22.65 & 0.7427 & 0.3693 & 1.80 & 28.72 & 0.8707 & 0.3264 & 1.81\\
 & 1/2 & 27.16 & 0.7635 & 0.3034 & 3.64 & 23.42 & 0.7817 & 0.3181 & 3.61 & 29.57 & 0.8913 & 0.2858 & 3.63\\
 & 3/4 & 27.59 & 0.7876 & 0.2679 & 5.46 & 23.61 & 0.7959 & 0.2933 & 5.41 & 29.69 & 0.8964 & 0.2706 & 5.44\\
 & Full & 27.89 & 0.8025 & 0.2414 & 8.41 & 23.71 & 0.8060 & 0.2708 & 8.43 & 29.76 & 0.8996 & 0.2552 & 8.76\\
\addlinespace[4pt]
\multirow{5}{*}{\shortstack[l]{S+O on\\\mbox{$A+C$}}} & 1/8 & 24.92 & 0.6603 & 0.4098 & 0.91 & 21.62 & 0.7085 & 0.4067 & 0.90 & 28.26 & 0.8617 & 0.3412 & 0.91\\
 & 1/4 & 26.54 & 0.7310 & 0.3415 & 1.82 & 23.02 & 0.7619 & 0.3480 & 1.80 & 29.28 & 0.8848 & 0.3000 & 1.81\\
 & 1/2 & 27.29 & 0.7720 & 0.2888 & 3.64 & 23.48 & 0.7898 & 0.3054 & 3.61 & 29.58 & 0.8948 & 0.2752 & 3.63\\
 & 3/4 & 27.62 & 0.7928 & 0.2577 & 5.46 & 23.60 & 0.8002 & 0.2847 & 5.41 & 29.65 & 0.8981 & 0.2626 & 5.44\\
 & Full & 27.76 & 0.8000 & 0.2446 & 7.28 & 23.64 & 0.8047 & 0.2739 & 7.21 & 29.68 & 0.8990 & 0.2567 & 7.26\\
\bottomrule
\end{tabular*}
\endgroup
\vspace{12pt}
}]

\newpage
\begin{nolinenumbers}
\par\smallskip\begingroup
\captionsetup{type=table,skip=4pt}
\centering\footnotesize
\setlength{\tabcolsep}{1.5pt}\renewcommand{\arraystretch}{1.18}
\renewcommand{\thetable}{S12}\caption{BD-rate (\%) for allocation and S+O. Allocation: \mbox{$A+C$} vs. $A$ with S+O off/on; S+O: on vs. off for $A$/\mbox{$A+C$}. Negative favors the first case \cite{bjontegaard,pchip}.}\label{tab:S12}
\begin{tabular*}{\columnwidth}{@{\extracolsep{\fill}}lrrrr@{}}
\toprule
\multirow{2}{*}{Dataset} & \multicolumn{2}{c}{Allocation} & \multicolumn{2}{c}{S+O}\\
\cmidrule(lr){2-3}\cmidrule(lr){4-5}
 & Off & On & $A$ & \mbox{$A+C$}\\\midrule
Mip-NeRF360 \cite{mipnerf360} & -22.00 & -19.84 & -14.87 & -16.19\\
Tanks \& Temples \cite{tanks} & -27.45 & -23.02 & -12.66 & -13.19\\
Deep Blending \cite{deepblending} & -36.29 & -31.53 & -9.97 & -9.61\\
\bottomrule\end{tabular*}
\par\endgroup
\end{nolinenumbers}


\clearpage
{\small\begin{thebibliography}{39}
\providecommand{\natexlab}[1]{#1}
\providecommand{\url}[1]{\texttt{#1}}
\expandafter\ifx\csname urlstyle\endcsname\relax
  \providecommand{\doi}[1]{doi: #1}\else
  \providecommand{\doi}{doi: \begingroup \urlstyle{rm}\Url}\fi

\bibitem[Barron et~al.(2022)Barron, Mildenhall, Verbin, Srinivasan, and
  Hedman]{mipnerf360}
Jonathan~T. Barron, Ben Mildenhall, Dor Verbin, Pratul~P. Srinivasan, and Peter
  Hedman.
\newblock {Mip-NeRF 360: Unbounded Anti-Aliased Neural Radiance Fields}.
\newblock In \emph{Proceedings of the IEEE/CVF Conference on Computer Vision
  and Pattern Recognition}, pages 5470--5479, 2022.

\bibitem[Chen et~al.(2026{\natexlab{a}})Chen, Chen, Zou, Huang, Wang, Liu, Sun,
  and Wang]{megs2}
Jiarui Chen, Yikeng Chen, Yingshuang Zou, Ye Huang, Peng Wang, Yuan Liu, Yujing
  Sun, and Wenping Wang.
\newblock {MEGS$^{2}$: Memory-Efficient Gaussian Splatting via Spherical
  Gaussians and Unified Pruning}.
\newblock In \emph{International Conference on Learning Representations},
  2026{\natexlab{a}}.

\bibitem[Chen et~al.(2025{\natexlab{a}})Chen, Wu, Li, Lin, Harandi, and
  Cai]{fcgs}
Yihang Chen, Qianyi Wu, Mengyao Li, Weiyao Lin, Mehrtash Harandi, and Jianfei
  Cai.
\newblock {Fast Feedforward 3D Gaussian Splatting Compression}.
\newblock In \emph{International Conference on Learning Representations}, pages
  74859--74872, 2025{\natexlab{a}}.

\bibitem[Chen et~al.(2025{\natexlab{b}})Chen, Wu, Lin, Harandi, and Cai]{hac}
Yihang Chen, Qianyi Wu, Weiyao Lin, Mehrtash Harandi, and Jianfei Cai.
\newblock {HAC: Hash-Grid Assisted Context for 3D Gaussian Splatting
  Compression}.
\newblock In \emph{Computer Vision {\textendash} ECCV 2024}, pages 422--438.
  Springer Nature Switzerland, 2025{\natexlab{b}}.

\bibitem[Chen et~al.(2025{\natexlab{c}})Chen, Wu, Lin, Harandi, and Cai]{hacpp}
Yihang Chen, Qianyi Wu, Weiyao Lin, Mehrtash Harandi, and Jianfei Cai.
\newblock {HAC++: Towards 100X Compression of 3D Gaussian Splatting}.
\newblock \emph{IEEE Transactions on Pattern Analysis and Machine
  Intelligence}, 47\penalty0 (11):\penalty0 10210--10226, 2025{\natexlab{c}}.

\bibitem[Chen et~al.(2026{\natexlab{b}})Chen, Li, Wu, Lin, Harandi, and
  Cai]{pcgs}
Yihang Chen, Mengyao Li, Qianyi Wu, Weiyao Lin, Mehrtash Harandi, and Jianfei
  Cai.
\newblock {PCGS: Progressive Compression of 3D Gaussian Splatting}.
\newblock In \emph{Proceedings of the AAAI Conference on Artificial
  Intelligence}, pages 3111--3119, 2026{\natexlab{b}}.

\bibitem[Di~Sario et~al.(2025)Di~Sario, Renzulli, Grangetto, Sugimoto, and
  Tartaglione]{gode}
Francesco Di~Sario, Riccardo Renzulli, Marco Grangetto, Akihiro Sugimoto, and
  Enzo Tartaglione.
\newblock {GoDe: Gaussians on Demand for Progressive Level of Detail and
  Scalable Compression}.
\newblock arXiv:2501.13558v2, 2025.

\bibitem[Fan et~al.(2024)Fan, Wang, Wen, Zhu, Xu, and Wang]{lightgaussian}
Zhiwen Fan, Kevin Wang, Kairun Wen, Zehao Zhu, Dejia Xu, and Zhangyang Wang.
\newblock {LightGaussian: Unbounded 3D Gaussian Compression with 15x Reduction
  and 200+ FPS}.
\newblock In \emph{Advances in Neural Information Processing Systems}, pages
  140138--140158, 2024.

\bibitem[Guo et~al.(2026)Guo, Zhang, Aktas, Fogarty, Hu, Aslan, Li, Baykal,
  Miao, Bengtson, Zhou, Xia, Vasconcelos, and Oztireli]{matryoshka}
Zhilin Guo, Boqiao Zhang, Hakan Aktas, Kyle Fogarty, Jeffrey Hu,
  Nursena~Koprucu Aslan, Wenzhao Li, Canberk Baykal, Albert Miao, Josef
  Bengtson, Chenliang Zhou, Weihao Xia, Cristina~Nader Vasconcelos, and Cengiz
  Oztireli.
\newblock {Matryoshka Gaussian Splatting}.
\newblock In \emph{Computer Vision {\textendash} ECCV 2026}, pages 323--340.
  Springer Nature Switzerland, 2026.

\bibitem[Hanson et~al.(2025)Hanson, Tu, Singla, Jayawardhana, Zwicker, and
  Goldstein]{pup}
Alex Hanson, Allen Tu, Vasu Singla, Mayuka Jayawardhana, Matthias Zwicker, and
  Tom Goldstein.
\newblock {PUP 3D-GS: Principled Uncertainty Pruning for 3D Gaussian
  Splatting}.
\newblock In \emph{Proceedings of the IEEE/CVF Conference on Computer Vision
  and Pattern Recognition}, pages 5949--5958, 2025.

\bibitem[Hedman et~al.(2018)Hedman, Philip, Price, Frahm, Drettakis, and
  Brostow]{deepblending}
Peter Hedman, Julien Philip, True Price, Jan-Michael Frahm, George Drettakis,
  and Gabriel Brostow.
\newblock {Deep blending for free-viewpoint image-based rendering}.
\newblock \emph{ACM Transactions on Graphics}, 37\penalty0 (6):\penalty0 1--15,
  2018.

\bibitem[Kerbl et~al.(2023)Kerbl, Kopanas, Leimk{\"u}hler, and
  Drettakis]{kerbl2023}
Bernhard Kerbl, Georgios Kopanas, Thomas Leimk{\"u}hler, and George Drettakis.
\newblock {3D Gaussian Splatting for Real-Time Radiance Field Rendering}.
\newblock \emph{ACM Transactions on Graphics}, 42\penalty0 (4):\penalty0 1--14,
  2023.

\bibitem[Knapitsch et~al.(2017)Knapitsch, Park, Zhou, and Koltun]{tanks}
Arno Knapitsch, Jaesik Park, Qian-Yi Zhou, and Vladlen Koltun.
\newblock {Tanks and Temples: Benchmarking Large-Scale Scene Reconstruction}.
\newblock \emph{ACM Transactions on Graphics}, 36\penalty0 (4):\penalty0 1--13,
  2017.

\bibitem[Koo et~al.(2026)Koo, Kim, and
  Ryu]{koo2026shinscalablehierarchyincremental}
Reagan Koo, Yeong-Gyu Kim, and Eun-Seok Ryu.
\newblock {SHIN: Scalable Hierarchy for INcremental 3D Gaussian Splatting}.
\newblock arXiv preprint arXiv:2611.10652, 2026.

\bibitem[Lee et~al.(2024)Lee, Rho, Sun, Ko, and Park]{compact3dgs}
Joo~Chan Lee, Daniel Rho, Xiangyu Sun, Jong~Hwan Ko, and Eunbyung Park.
\newblock {Compact 3D Gaussian Representation for Radiance Field}.
\newblock In \emph{Proceedings of the IEEE/CVF Conference on Computer Vision
  and Pattern Recognition}, pages 21719--21728, 2024.

\bibitem[Lee et~al.(2025{\natexlab{a}})Lee, Ko, and Park]{omg}
Joo~Chan Lee, Jong~Hwan Ko, and Eunbyung Park.
\newblock {Optimized Minimal 3D Gaussian Splatting}.
\newblock In \emph{Advances in Neural Information Processing Systems}, pages
  135864--135888, 2025{\natexlab{a}}.

\bibitem[Lee et~al.(2025{\natexlab{b}})Lee, Shu, Sanchez, Schierl, and
  Hellge]{featureplanes2025}
Soonbin Lee, Fangwen Shu, Yago Sanchez, Thomas Schierl, and Cornelius Hellge.
\newblock Compression of {3D Gaussian Splatting} with optimized feature planes
  and standard video codecs.
\newblock In \emph{Proceedings of the IEEE/CVF International Conference on
  Computer Vision}, pages 25496--25505, 2025{\natexlab{b}}.

\bibitem[Liu et~al.(2025{\natexlab{a}})Liu, Wang, Li, Cai, Wang, Li, Molchanov,
  Wang, and Wang]{flexgs}
Hengyu Liu, Yuehao Wang, Chenxin Li, Ruisi Cai, Kevin Wang, Wuyang Li, Pavlo
  Molchanov, Peihao Wang, and Zhangyang Wang.
\newblock {FlexGS: Train Once, Deploy Everywhere with Many-in-One Flexible 3D
  Gaussian Splatting}.
\newblock In \emph{Proceedings of the IEEE/CVF Conference on Computer Vision
  and Pattern Recognition}, pages 16336--16345, 2025{\natexlab{a}}.

\bibitem[Liu et~al.(2025{\natexlab{b}})Liu, Zhong, Zhan, Xu, and
  Sun]{maskgaussian}
Yifei Liu, Zhihang Zhong, Yifan Zhan, Sheng Xu, and Xiao Sun.
\newblock {MaskGaussian: Adaptive 3D Gaussian Representation from Probabilistic
  Masks}.
\newblock In \emph{Proceedings of the IEEE/CVF Conference on Computer Vision
  and Pattern Recognition}, pages 681--690, 2025{\natexlab{b}}.

\bibitem[Milef et~al.(2025)Milef, Seyb, Keeler, Nguyen-Phuoc, Bo{\v z}i{\v c},
  Kondguli, and Marshall]{clod}
Nicholas Milef, Dario Seyb, Todd Keeler, Thu Nguyen-Phuoc, Alja{\v z} Bo{\v
  z}i{\v c}, Sushant Kondguli, and Carl Marshall.
\newblock {Learning Fast 3D Gaussian Splatting Rendering using Continuous Level
  of Detail}.
\newblock \emph{Computer Graphics Forum}, 44\penalty0 (2):\penalty0 e70069,
  2025.

\bibitem[Morgenstern et~al.(2026)Morgenstern, Branschke, Fleischmann, Szatmari,
  Schlack, Barthel, Eisert, and Hilsmann]{kissgs}
Wieland Morgenstern, Friedrich~Elias Branschke, Florian Fleischmann, Adrian
  Szatmari, Paul Schlack, Florian Barthel, Peter Eisert, and Anna Hilsmann.
\newblock {KISS-GS: 3D Gaussian Splatting Compression Kept Simple}.
\newblock In \emph{Computer Vision {\textendash} ECCV 2026}, pages 677--694.
  Springer Nature Switzerland, 2026.

\bibitem[Navaneet et~al.(2025)Navaneet, Pourahmadi~Meibodi,
  Abbasi~Koohpayegani, and Pirsiavash]{compgs}
K.~L. Navaneet, Kossar Pourahmadi~Meibodi, Soroush Abbasi~Koohpayegani, and
  Hamed Pirsiavash.
\newblock {CompGS: Smaller and Faster Gaussian Splatting with Vector
  Quantization}.
\newblock In \emph{Computer Vision {\textendash} ECCV 2024}, pages 330--349.
  Springer Nature Switzerland, 2025.

\bibitem[Niedermayr et~al.(2024)Niedermayr, Stumpfegger, and
  Westermann]{compressed3dgs}
Simon Niedermayr, Josef Stumpfegger, and R\"udiger Westermann.
\newblock {Compressed 3D Gaussian Splatting for Accelerated Novel View
  Synthesis}.
\newblock In \emph{Proceedings of the IEEE/CVF Conference on Computer Vision
  and Pattern Recognition}, pages 10349--10358, 2024.

\bibitem[Pan et~al.(2026)Pan, Zhou, and Zha]{ldh}
Youqi Pan, Wugen Zhou, and Hongbin Zha.
\newblock {Learning Differentiable Hierarchies in 3D Gaussian Splatting}.
\newblock In \emph{Proceedings of the IEEE/CVF Conference on Computer Vision
  and Pattern Recognition (CVPR)}, pages 40939--40948, 2026.

\bibitem[Ren et~al.(2025)Ren, Jiang, Lu, Yu, Xu, Ni, and Dai]{octreegs}
Kerui Ren, Lihan Jiang, Tao Lu, Mulin Yu, Linning Xu, Zhangkai Ni, and Bo Dai.
\newblock {Octree-GS: Towards Consistent Real-time Rendering with
  LOD-Structured 3D Gaussians}.
\newblock \emph{IEEE Transactions on Pattern Analysis and Machine
  Intelligence}, pages 1--15, 2025.

\bibitem[Revilla et~al.(2026)Revilla, Suresh, Tsang, and Bhojan]{scargs}
Diego Revilla, Pooja Suresh, Ooi~Wei Tsang, and Anand Bhojan.
\newblock {SCAR-GS: Spatial Context Attention for Residuals in Progressive
  Gaussian Splatting}.
\newblock arXiv:2601.04348v2, 2026.

\bibitem[Seo et~al.(2025)Seo, Choi, Son, and Uh]{flod}
Yunji Seo, Young~Sun Choi, HyunSeung Son, and Youngjung Uh.
\newblock {FLoD: Integrating Flexible Level of Detail into 3D Gaussian
  Splatting for Customizable Rendering}.
\newblock \emph{ACM Transactions on Graphics}, 44\penalty0 (4):\penalty0 1--16,
  2025.

\bibitem[Shi et~al.(2025)Shi, Morin, Gasparini, and Ooi]{lapis}
Yuang Shi, G\'eraldine Morin, Simone Gasparini, and Wei~Tsang Ooi.
\newblock {LapisGS: Layered Progressive 3D Gaussian Splatting for Adaptive
  Streaming}.
\newblock In \emph{2025 International Conference on 3D Vision (3DV)}, pages
  991--1000, 2025.

\bibitem[Tang et~al.(2026)Tang, Liu, Jiao, Zhang, Yang, Hou, and Wang]{progs}
Zhiye Tang, Lingzhuo Liu, Shengjie Jiao, Qiudan Zhang, You Yang, Junhui Hou,
  and Xu Wang.
\newblock {ProGS: Towards Progressive Coding for 3D Gaussian Splatting}.
\newblock arXiv:2603.09703v2, 2026.

\bibitem[Tian et~al.(2025)Tian, Gao, Xianyu, Cui, and Zhang]{flexgaussian}
Boyuan Tian, Qizhe Gao, Siran Xianyu, Xiaotong Cui, and Minjia Zhang.
\newblock {FlexGaussian: Flexible and Cost-Effective Training-Free Compression
  for 3D Gaussian Splatting}.
\newblock In \emph{Proceedings of the 33rd ACM International Conference on
  Multimedia}, pages 7287--7296, 2025.

\bibitem[Wang et~al.(2025)Wang, Zhu, He, Feng, Deng, Bian, and Chen]{rdo}
Henan Wang, Hanxin Zhu, Tianyu He, Runsen Feng, Jiajun Deng, Jiang Bian, and
  Zhibo Chen.
\newblock {End-to-End Rate-Distortion Optimized 3D Gaussian Representation}.
\newblock In \emph{Computer Vision {\textendash} ECCV 2024}, pages 76--92.
  Springer Nature Switzerland, 2025.

\bibitem[Wang et~al.(2026)Wang, Zeng, and Fu]{wang2026racergs}
Wenqing Wang, Huimin Zeng, and Yun Fu.
\newblock {RacerGS: Residual-Autoregressive Context for 3D Gaussian Splatting
  Compression}.
\newblock In \emph{Advances in Neural Information Processing Systems}, 2026.

\bibitem[Wang et~al.(2024)Wang, Li, Guo, Yang, Kot, and Wen]{contextgs}
Yufei Wang, Zhihao Li, Lanqing Guo, Wenhan Yang, Alex~C. Kot, and Bihan Wen.
\newblock {ContextGS : Compact 3D Gaussian Splatting with Anchor Level Context
  Model}.
\newblock In \emph{Advances in Neural Information Processing Systems}, pages
  51532--51551, 2024.

\bibitem[Xie et~al.(2025{\natexlab{a}})Xie, Liu, Zhang, Ge, Pan, Tang, Bai,
  Zhang, Fan, and Wang]{sizegs}
Shuzhao Xie, Jiahang Liu, Weixiang Zhang, Shijia Ge, Sicheng Pan, Chen Tang,
  Yunpeng Bai, Cong Zhang, Xiaoyi Fan, and Zhi Wang.
\newblock {SizeGS: Size-aware Compression of 3D Gaussian Splatting via Mixed
  Integer Programming}.
\newblock In \emph{Proceedings of the 33rd ACM International Conference on
  Multimedia}, pages 8214--8223, 2025{\natexlab{a}}.

\bibitem[Xie et~al.(2025{\natexlab{b}})Xie, Zhang, Tang, Bai, Lu, Ge, and
  Wang]{mesongs}
Shuzhao Xie, Weixiang Zhang, Chen Tang, Yunpeng Bai, Rongwei Lu, Shijia Ge, and
  Zhi Wang.
\newblock {MesonGS: Post-training Compression of 3D Gaussians via Efficient
  Attribute Transformation}.
\newblock In \emph{Computer Vision {\textendash} ECCV 2024}, pages 434--452.
  Springer Nature Switzerland, 2025{\natexlab{b}}.

\bibitem[Yu et~al.(2026)Yu, Chen, Song, Qin, Zhang, Wang, and
  Guo]{yu2026practicalcompression3dgaussian}
Pengpeng Yu, Yueru Chen, Fei Song, Tai Qin, Qi Zhang, Jing Wang, and Yulan Guo.
\newblock {Towards Practical Compression of 3D Gaussian Splatting}.
\newblock arXiv:2609.30245, 2026.

\bibitem[Zhang et~al.(2025)Zhang, Jia, Niu, and Yin]{gaussianspa}
Yangming Zhang, Wenqi Jia, Wei Niu, and Miao Yin.
\newblock {GaussianSpa: An
  {\textquotedblleft}Optimizing-Sparsifying{\textquotedblright} Simplification
  Framework for Compact and High-Quality 3D Gaussian Splatting}.
\newblock In \emph{Proceedings of the IEEE/CVF Conference on Computer Vision
  and Pattern Recognition}, pages 26673--26682, 2025.

\bibitem[Zheng et~al.(2025)Zheng, Wu, Zhong, Tian, Cao, Xu, Yao, Zhang, Hu, and
  Zhang]{4dgcpro}
Zihan Zheng, Zhenlong Wu, Houqiang Zhong, Yuan Tian, Ning Cao, Lan Xu,
  Jiangchao Yao, Xiaoyun Zhang, Qiang Hu, and Wenjun Zhang.
\newblock {4DGCPro}: Efficient hierarchical {4D Gaussian} compression for
  progressive volumetric video streaming.
\newblock In \emph{Advances in Neural Information Processing Systems}, pages
  100860--100885, 2025.

\bibitem[Zoomers et~al.(2025)Zoomers, Wijnants, Molenaers, Vanherck, Put,
  Jorissen, and Michiels]{progs_wacv}
Brent Zoomers, Maarten Wijnants, Ivan Molenaers, Joni Vanherck, Jeroen Put,
  Lode Jorissen, and Nick Michiels.
\newblock {PRoGS: Progressive Rendering of Gaussian Splats}.
\newblock In \emph{Proceedings of the Winter Conference on Applications of
  Computer Vision}, pages 3118--3127, 2025.

\end{thebibliography}
}


\clearpage
{\small\begin{thebibliography}{43}
\providecommand{\natexlab}[1]{#1}
\providecommand{\url}[1]{\texttt{#1}}
\expandafter\ifx\csname urlstyle\endcsname\relax
  \providecommand{\doi}[1]{doi: #1}\else
  \providecommand{\doi}{doi: \begingroup \urlstyle{rm}\Url}\fi

\setcounter{NAT@ctr}{39}

\bibitem[Bjontegaard(2001)]{bjontegaard}
Gisle Bjontegaard.
\newblock {Calculation of average PSNR differences between RD-curves}.
\newblock Technical Report VCEG-M33, ITU-T SG16/Q6 Video Coding Experts Group,
  2001.

\bibitem[Fritsch and Butland(1984)]{pchip}
F.~N. Fritsch and J. Butland.
\newblock {A Method for Constructing Local Monotone Piecewise Cubic
  Interpolants}.
\newblock \emph{SIAM Journal on Scientific and Statistical Computing},
  5\penalty0 (2):\penalty0 300--304, 1984.

\bibitem[Krichevsky and Trofimov(1981)]{kt}
R. Krichevsky and V. Trofimov.
\newblock {The performance of universal encoding}.
\newblock \emph{IEEE Transactions on Information Theory}, 27\penalty0
  (2):\penalty0 199--207, 1981.

\bibitem[Witten et~al.(1987)Witten, Neal, and Cleary]{arithmetic}
Ian~H. Witten, Radford~M. Neal, and John~G. Cleary.
\newblock {Arithmetic coding for data compression}.
\newblock \emph{Communications of the ACM}, 30\penalty0 (6):\penalty0 520--540,
  1987.

\end{thebibliography}
}
\end{document}